\documentclass{article}
\usepackage{placeins}
\usepackage{float}
\usepackage{adjustbox}
\usepackage{makecell}
\usepackage{multirow}
\usepackage{amsmath}
    \PassOptionsToPackage{numbers, compress}{natbib}
 \usepackage[preprint]{neurips_2026}

\usepackage{hyperref}       
\usepackage{url}            
\usepackage{booktabs}       
\usepackage{amsfonts}       
\usepackage{nicefrac}       
\usepackage{microtype}      
\usepackage{xcolor}         
\usepackage{placeins}
\usepackage{cleveref}
\usepackage{float}
\usepackage{enumitem}

\usepackage{xspace}
\newcommand{\layerscope}{\textsc{LayerScope}\xspace}

\title{\layerscope: A Layerwise Characterization of Video and Multimodal Learned Representations}

\usepackage{xcolor}

\newcommand{\newtodo}[3]{%
  \expandafter\newcommand\csname #1\endcsname[1]{%
    \textcolor{#3}{\textbf{TODO (#2): ##1}}%
  }%
}

\author{%
Sandra Arcos-Holzinger\thanks{These authors contributed equally.}$^{*1,4}$ \quad Debashish Chakraborty\footnotemark[1]$^{*2}$ \quad 
Rohita Mocharla\footnotemark[1]$^{*1,3}$ \\
\textbf{Will Walden}$^{1,2}$\quad 
\textbf{Andrew Yates}$^{1,2}$\quad 
\textbf{Reno Kriz}$^{1,2}$\\
\textbf{Sarah M. Erfani}$^4$ \quad \textbf{James Bailey}$^5$ \quad 
\textbf{Vishal M. Patel}$^{1}$ \quad \textbf{Sanjeev Khudanpur}$^{1,2}$\\
$^1$Johns Hopkins University \quad $^2$Human Language Technology Center of Excellence \\
$^3$Johns Hopkins Applied Physics Lab \quad $^4$University of Melbourne \quad $^5$Monash University \\
\texttt{\{sarcosh1,dchakra6,nmochar1\}@jh.edu}
}

\begin{document}

\maketitle
\begin{abstract}
We propose \layerscope, a label-free, layerwise framework that aims to characterize a model's learned representations in video and multimodal settings. Evaluating downstream performance using representations from final or intermediate layers typically requires large amounts of labeled data, repeated task-specific evaluations, and substantial computation. To address these limitations, \layerscope uses local, global, distributional, and correspondence-based geometric metrics to compare layerwise representation structure within and across models without requiring task-specific labels. We evaluate seven architecturally diverse models across video and multimodal classification, clustering, and text-to-video retrieval tasks from MVEB/MVEB+. We find that intermediate-layer representations can outperform final-layer and model-default outputs. We also find that no single geometric metric consistently predicts downstream performance, but note that distinct layerwise geometric signatures emerge across model families. LID shows \textit{task-dependent} relationships with performance, while RankMe provides the strongest measure for classification and clustering, but is not a universal layer selector. We also find that pairing-aware metrics explain retrieval better than distributional distances alone. \layerscope therefore offers a framework for comparing representations across models and layers, enabling a more systematic evaluation in video and multimodal settings.
\end{abstract}
\section{Introduction}
\label{intro}

Multimodal embedding models are typically evaluated through downstream benchmarks  such as MVEB~\cite{assadi2026mvebmassivevideoembedding} or MTEB~\cite{muennighoff2023mtebmassivetextembedding} and their default output representations. These benchmarks measure capability across tasks including classification, clustering, and retrieval, but treat the internal learned representations as a black box. Thus, benchmark scores merely indicate model success or failure, but provide limited insight into how useful representations develop throughout the network. Evaluating intermediate layers can be computationally expensive and task-dependent. Geometric measures provide \textit{label-free, task-independent diagnostics} of representation structure: the metrics themselves do not depend on downstream labels or task scores, although their relationship with downstream performance may vary across tasks. While geometric measures do not replace benchmark performance, they characterize the intrinsic properties of the learned representations i.e., local, global, distributional and correspondence-sensitive properties that downstream scores do not expose.

In this work, we refer to \textit{label-free }geometric diagnostics for metrics that do not require downstream labels or task scores, and are primarily computed using a model's learned representations. We also refer to \textit{task-dependent} conditions, in cases where a model's geometric properties changes with downstream performance. Our work aims to investigate beyond \textit{whether} geometric measures capture \textit{representation quality}, but to identify the conditions \textit{under which} geometric metrics may explain, characterize, or predict useful downstream behavior in models.

\paragraph{Why layerwise representation geometry may help.}
Internal representations can change substantially across layers in the model, and useful semantic information can emerge before the final layer in a network; this motivates auditing intermediate representations rather than considering only the output of the last layer as the object of evaluation~\cite{bolya2025PerceptionEncoder}. This has direct implications across encoder model families and objective functions used during training. Geometric quantities such as local intrinsic dimensionality (LID), effective rank~\cite{roy2007effectiverank}, and cross-modal distributional distance provide different mathematical approaches to characterize the high dimensional structure of a model's representations---and in some cases, without requiring task labels. Global and local geometric diagnostics provide complementary views of learned representations. Global measures characterize dataset-level dimensionality or spectral utilization, whereas local measures such as LID characterize neighborhood-level manifold complexity. Prior work such as GRIDS-LID \cite{arcosholzinger2026dimensionalityawareanomalydetectionlearned} motivates intrinsic-dimensionality analysis of learned representations for anomaly detection in self-supervised speech models (S3Ms), but it remains unclear whether intrinsic metrics such as LID are reliable diagnostics for the evaluation of different model architectures, layers, modalities, and downstream tasks. In this context, \layerscope  asks whether geometric metrics can explain or predict downstream performance by inspection of a model's representations. Geometry is thus not a replacement for representation quality, but rather a complementary metric to characterize learned representations and capture when geometric diagnostics agree with, or diverge from downstream performance. Positive results establish where geometry is informative; negative results  pinpoint to the limits of the diagnostics, e.g.,  constraints in LID as a single metric when identifying the best-performing layer.

\begin{figure}[t]
    \centering
    \includegraphics[width=1.0\linewidth, height=8cm, keepaspectratio
    ]{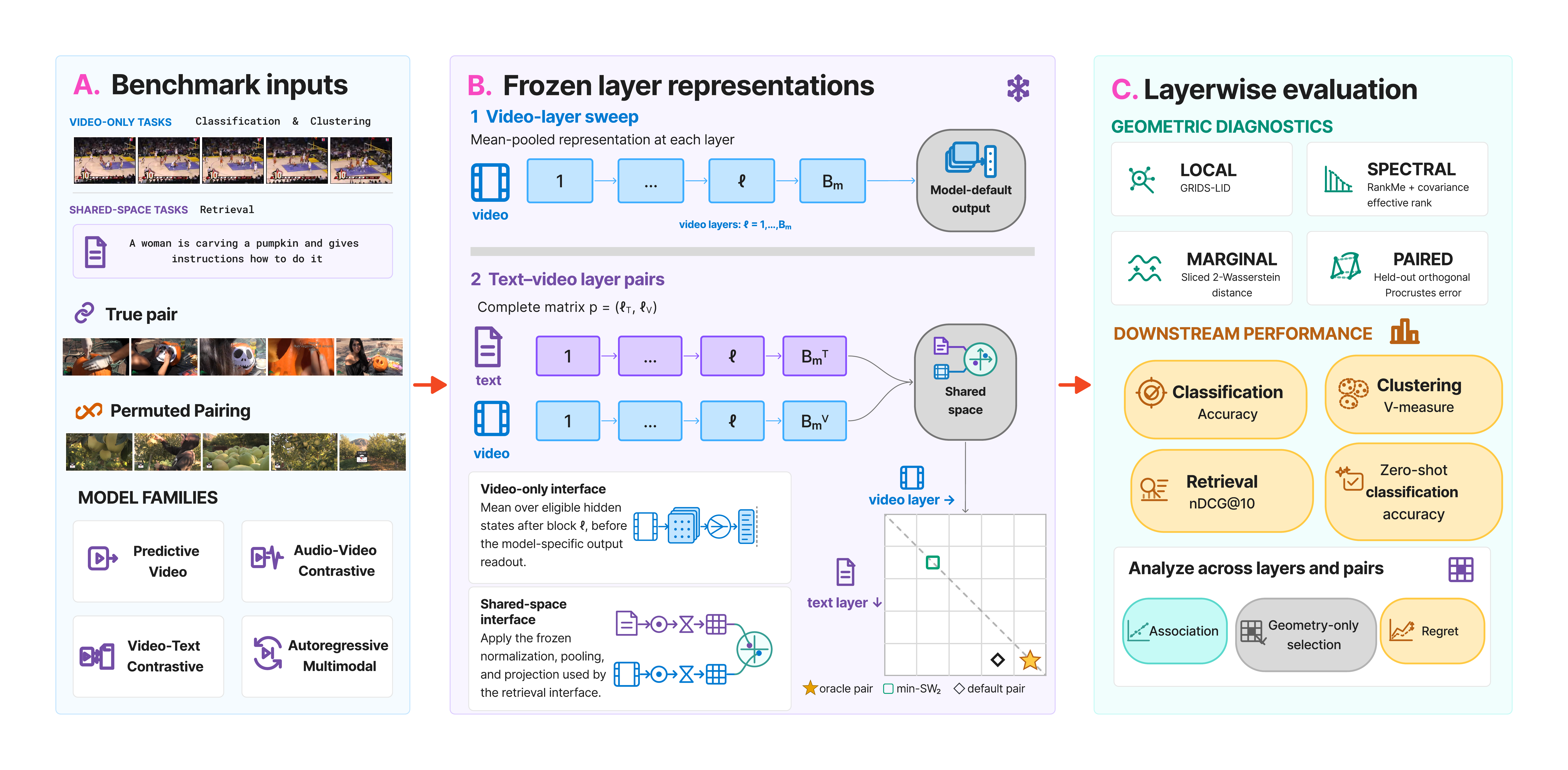}
    \caption{\layerscope framework. Given video-only tasks or tasks evaluated in a shared text–video space, \layerscope freezes each model and evaluates representations across transformer depth. For classification and clustering, eligible post-block hidden states are mean pooled; model-default outputs are evaluated separately. For models with shared text–video spaces, the model-specific normalization, pooling, and projection operations are applied at every text and video layer, producing the complete pair matrix \(p=(\ell_T,\ell_V)\in\mathcal P_{m,t}\). 
    \layerscope relates GRIDS-LID, RankMe, covariance effective rank, sliced 2-Wasserstein distance, and held-out orthogonal Procrustes error to downstream performance through within-model association and geometry-only selection regret. 
    }
    \label{fig:study_map}
\end{figure}

\paragraph{Contributions.}
We study how the geometry and downstream performance of video and multimodal representations evolve across network depth. To do so, we introduce \layerscope, a layerwise evaluation framework that jointly audits intermediate representations using downstream tasks and complementary local, global, and cross-modal geometric diagnostics. This allows us to address three distinct questions: \textit{first}, how representation geometry evolves across network depth; \textit{second}, whether geometric measures are associated with downstream performance; and \textit{third}, whether those measures can reliably identify high-performing representations.

Using \layerscope, we first ask when intermediate representations outperform final-layer or usual model outputs across classification, clustering, zero-shot classification, and text-to-video retrieval (T2VR). We then examine whether intrinsic geometric metrics track layerwise downstream performance, and whether layer selection using these metrics can identify high-performing representations with low regret. Finally, for multimodal models, we study how text and video representations relate across network depth; this is achieved by distinguishing marginal distributional proximity from pair-sensitive alignment and evaluating how each relates to retrieval behavior. Across predictive, contrastive, and autoregressive model families, we also compare their layerwise geometric profiles, allowing us to understand the evolution of representations in the network and emerging patterns which we refer to as \textit{geometric signatures.}

\layerscope evaluates models using geometric diagnostics to uncover aspects of representation structure that are not directly observable from benchmark scores alone. Our framework does not propose geometric analyses as alternatives to downstream tasks, but rather, as complementary \textit{label-free} evaluation proxies. In our work, these support the evaluation of failure modes in a model, as well as layer-selection criteria. \Cref{fig:study_map} summarizes the framework and experimental workflow.

Our contributions are:
\begin{enumerate}[label=(\roman*)]
    \item 
    We introduce \layerscope, a layerwise evaluation framework that jointly audits downstream performance and local, global, and cross-modal representation geometry across seven heterogeneous video and multimodal models;

    \item
    We conduct a systematic layerwise evaluation across classification, clustering, zero-shot classification, and T2VR, and analyze geometry as both a performance correlate and a \textit{label-free }layer-selection criterion using association and selection regret; and

    \item
    We distinguish marginal distributional proximity from pair-sensitive alignment, showing when geometric measures track retrieval behavior and when correspondence-aware diagnostics reveal text--video structure that marginal distribution measures cannot capture.
\end{enumerate}
\section{Related Work}
\label{related_work}

\paragraph{Embedding and multimodal evaluation benchmarks.}
Embedding benchmarks have expanded from individual modalities toward broad, task-diverse evaluation suites. MTEB standardized evaluation of text embeddings across multiple downstream tasks~\cite{muennighoff2023mtebmassivetextembedding}, while MSEB extended this style of evaluation to sound representations~\cite{heigold2026massivesoundembeddingbenchmark}. More recently, MVEB introduced a video embedding benchmark spanning (zero-shot) classification, pair classification, clustering, retrieval, and video-centric question answering---with MVEB+ providing a larger pool of tasks across video, text, and audio~\cite{assadi2026mvebmassivevideoembedding}. These benchmarks measure downstream performance, but do not generally examine how learned representations vary  \textit{within} the network, nor do they commonly treat network depth as an evaluation variable i.e., layer number. For example, MVEB evaluates models using its model-default embedding output without layer-specific tuning~\cite{assadi2026mvebmassivevideoembedding}. This leaves a few open questions: how does downstream performance vary across layers in a network? Is it possible to identify high-performing internal representations in a model, without exhaustively evaluating every layer on labeled tasks? From this perspective, \layerscope complements downstream benchmarking by treating network-depth (i.e., number of hidden layers) as an evaluation variable, and testing whether \textit{label-free} layerwise geometric measures can be both descriptive and predictive of downstream behavior.

\paragraph{Geometry of learned representations.} 
The geometric view of representation learning is motivated by the manifold hypothesis, under which high-dimensional observations can concentrate near a lower-dimensional structure \cite{NIPS2010_8a1e808b,fefferman2016testing}. Studies of neural representations similarly find that learned features can occupy low-dimensional structures within much larger ambient spaces \cite{lid-Ansuini2019,lid-Gong2019,lid-Pope2021}, and recent work has examined how computations inside transformers can modify manifold structure across layers \cite{gurnee2026modelsmanipulatemanifoldsgeometry_anthropic}. These results motivate intrinsic and spectral measurements as tools for studying how representation structure changes throughout a network. This includes complementary local and global measurements that aid in the analysis of latent representations. Respectively, at the local scale, metrics are more granular in capturing neighborhood structure, while at the global scale, metrics are more coarse and use the full embedding space. These approaches have been used extensively in data mining \cite{Bailey2022LocalID} with clear applications for distributional analysis and outlier detection of latent representations.

For example, global rank-based methods summarize global spectral structure and have been proposed as \textit{label-free} indicators associated with downstream performance~\cite{rankme-garrido-lecun-2023}. In S3Ms, global rank measures have been found to correlate with downstream performance, although they do not reliably identify the best-performing layer for a given task~\cite{aldeneh2024rankme,rank-Whetten2025}. At the local scale, LID provides a complementary view by estimating the effective dimensionality of the neighborhood around a query point (e.g., embedding corresponding to an image frame) in relation to nearest-neighbor distances; thus, measuring the rate at which the mass of such neighborhood grows with distance \cite{lid-amsaleg1028,lid-bailey2021,NIPS2004_74934548}. LID has been used to characterize adversarial examples, noisy-label memorization, and contextual-language-model training dynamics \cite{ma2018characterizing,ma2018dimensionality,ruppik2025less}. Related local density statistics have been used to identify backdoor-poisoned samples \cite{huang2025detectingbackdoorsamplescontrastive-localsubspace-LID-JamesB}, while GRIDS-LID uses layerwise LID features for anomaly detection in S3Ms \cite{arcosholzinger2026dimensionalityawareanomalydetectionlearned}. 

In the context of recent applications, LID is motivated as an evaluation metric of local structure of latent representations; however, it does not imply that a high or low LID value is a universal indicator of better downstream performance. GRIDS-LID \cite{arcosholzinger2026dimensionalityawareanomalydetectionlearned} builds on this distinction by using LID as a layerwise local geometric diagnostic while contrasting it with global dimensionality and similarity-based approaches. A key distinction is that intrinsic dimensionality is not equivalent to representation quality. LID describes local geometric complexity, while effective rank describes global spectral utilization, and the relationship between either quantity and downstream performance remains empirical. Therefore, \layerscope evaluates these measurements against classification, clustering, and retrieval behavior across architectures and network-depth rather than assuming that a particular geometric measurement is universal indicator of performance.

\paragraph{Intermediate representations and layer selection}
Intermediate representations can retain task-relevant information that is weakened or specialized in the final layers, making network-depth an important downstream evaluation. This pattern has been observed in vision encoders \cite{bolya2025PerceptionEncoder} and in layerwise studies of self-supervised speech models \cite{Pasad_2021-cca-layerwise-Livescu,pasad2023comparative}. Representation-similarity metrics such as centered kernel alignment (CKA) quantify similarity between representations spaces across layers in a model ~\cite{kornblith2019similarity-cka-Hinton}. Recent work has also studied intermediate language model representations across text and vision tasks \cite{skean2025layer}, selected task-discriminative subspaces with spectral criteria \cite{batra2026loes}, and combined intermediate video-MLLM states with a calibrated retrieval head \cite{tzachor2026vidvec}. Our study instead examines a set of \textit{label-free} geometric diagnostics across classification, clustering, and complete text-video layer configurations. This allows to distinguish association with downstream performance, layer selection, and cross-modal correspondence, rather than treating these as a single aspect of layerwise representation analysis.

\paragraph{Cross-modal alignment and distributional proximity.} 
Multimodal models are often trained so that semantically related inputs from different modalities can be matched in a shared representation space, yet successful matching does not imply that different modalities will have identical distributions.
Prior work on contrastive multimodal learning has shown the presence of modality gaps both under random and pretrained weight initialization \cite{ModalityGap}. Other work has also shown that reducing this gap can improve retrieval in some settings \cite{yaras2025explaining}, harm performance when modality-specific information is removed \cite{jiang2023understanding}, or impact robustness and semantics at different levels when comparing group-level retrieval versus instance-level retrieval \cite{chowers2026bug,grassucci2026closing}.
Together, these results motivate treating distributional proximity - rather than a strict distributional identity - as the key representational property underlying semantic matching in retrieval. As such, \layerscope aims to study this relationship between proximity and performance across transformer layers in a model. Analyses of intermediate vision-language representations also suggest that cross-modal integration develops progressively across layers \cite{huang2024deciphering,Song_2026_CVPR}, while large-scale benchmarks such as MVEB \cite{assadi2026mvebmassivevideoembedding} show that retrieval behavior varies substantially across multimodal model families and modality combinations.

In our work, we therefore make a clear distinction regarding \textit{distributional proximity}: the closeness of the overall text and video representation distributions; \textit{cross-modal correspondence}: whether the correct text and video examples are associated with one another; and \textit{semantic alignment}: whether corresponding inputs expressing the same underlying semantic content can be correctly matched. \layerscope, also aims to study retrieval behavior through sliced Wasserstein distance \cite{nietert2022slicedwasserstein} and Procrustes error \cite{Schonemann1966-orthogonal-procrustes} \cite{Grave2018-wasserstein-procrustes}. These metrics serve as measures of representation alignment; the first evaluating marginal text-video distributional proximity, while the second determining paired cross-modal correspondence.
\section{Methods} 
\label{method}

We evaluate the hidden representations of seven video and multimodal models on classification, clustering, and T2VR. For each model and dataset, we extract the layerwise representations once, compute geometric measurements from the resulting cache, and evaluate the same representations on the corresponding downstream task. We compare geometry with performance across depth and use regret to quantify the cost of choosing a layer from geometry alone.

\subsection{Problem setting} \label{sec:problem_setting}
For model \(m\) and task \(t\), let
\(\mathcal{B}_{m,t}\subseteq\{1,\ldots,B_m\}\) denote the set of evaluated layers for which geometry and downstream performance are available.
Let \(\widehat{\ell}_{g}\in\mathcal{B}_{m,t}\) be the layer selected by geometric measure \(g\) without consulting downstream scores. 
For evaluation seed \(s\), let \[\ell^{\star}_{m,t,s}=\arg\max_{\ell\in\mathcal{B}_{m,t}}P_{m,t,s}(\ell)
\] denote the oracle best-performing layer, where \(P_{m,t,s}(\ell)\) is performance at layer \(\ell\) on task $t$ for evaluation seed \(s\). Following \cite{renggli2022which}, we report absolute regret:

\begin{equation}
R_{m,t,s}(g)
=
P_{m,t,s}(\ell^{\star}_{m,t,s})
-
P_{m,t,s}(\widehat{\ell}_{g}),
\label{eq:regret}
\end{equation}

For retrieval, let \(\mathcal{P}_{m,t}\) denote the set of evaluated
text-video layer pairs \(p=(\ell_T,\ell_V)\). Retrieval is deterministic
for fixed representations, and so is not indexed by evaluation seed. We define the
retrieval oracle as
\[
p^\star_{m,t}
=
\arg\max_{p\in\mathcal{P}_{m,t}} P_{m,t}(p),
\]
where \(P_{m,t}(p)\) is mean nDCG@10 for pair \(p\). If geometric measure \(g\) selects the set
\(\widehat{\mathcal{P}}_{m,t}(g)\subseteq\mathcal{P}_{m,t}\),
the pair-selection regret is
\begin{equation}
R^{\mathrm{pair}}_{m,t}(g)
=
P_{m,t}(p^\star_{m,t})
-
\max_{p\in\widehat{\mathcal{P}}_{m,t}(g)}
P_{m,t}(p).
\label{eq:pair_regret}
\end{equation}

For the minimum sliced 2-Wasserstein distance (SW$_2$) selector \cite{nietert2022slicedwasserstein}, \(\widehat{\mathcal{P}}_{m,t}(\mathrm{SW_2})\) contains all exact minimum-SW$_2$ pairs. We use the same oracle-minus-reference convention for the model-default pair and matched-depth baseline, but report these separately from geometry-based selection.

\subsection{Models, representations, and tasks} \label{sec:models_representations_tasks} 

We treat the final-layer representation and model-default output as distinct representations.
We define the \textit{model-default output} as the embedding after any additional normalization, pooling, projection, etc. For a model \(m\) with \(B_m\) transformer blocks, layer \(\ell \in \{1,\ldots,B_m\}\) refers to the hidden state immediately after transformer block \(\ell\). On the other hand, we refer to the representation obtained from the hidden state at \(\ell=B_m\) as the \textit{final-layer representation}.

\paragraph{Models.} 
We evaluate V-JEPA 2 ViT-L \cite{assran2025vjepa2}, PE-AV Small and Large \cite{vyas2025pushingfrontieraudiovisualperception}, Gemma 4 E4B-IT \cite{googledeepmind2026gemma4}, X-CLIP Base \cite{ni2022expanding}, Qwen3-VL-Embedding-2B \cite{li2026qwen3vlembeddingqwen3vlrerankerunifiedframework}, and LCO-Embedding-Omni-3B~\cite{xiao2025scaling}. 
All model parameters remain frozen, and we train no layer-specific or cross-modal alignment adapters.
The models differ in architecture, training data, objective, and model-default output, so we treat layer profiles as descriptive comparisons rather than strictly caused by difference in training objectives. Model-specific representation extraction and checkpoints are listed in \Cref{tab:model_inventory}.

\paragraph{Representations.} 
For classification and clustering, the representation at layer $\ell$ is the mean of the hidden states after transformer block \(\ell\) and before the model-default output. We exclude padding and special token hidden states. For LCO-Embedding-Omni (3B) (LCO), the mean is taken over eligible modality token hidden states. 
In order to ensure layer representation in the model's shared text-video embedding space for retrieval tasks, we apply model-specific normalization, pooling, and projection.
For a single representation stream, we define normalized depth is defined to be \(\tau_\ell=\ell/B_m\). For classification and clustering, we evaluate every video layer in all seven models. Retrieval evaluates every (text layer, video layer) pair for the five models with a shared text-video space.

\paragraph{Tasks and evaluation.} Task definitions follow MVEB and MVEB+ through MTEB \cite{assadi2026mvebmassivevideoembedding,muennighoff2023mtebmassivetextembedding}. We evaluate Breakfast classification \cite{kuehne2014language}, UCF101 classification and clustering \cite{soomro2012ucf101}, and T2VR on TUNA-Bench \cite{kong2025tuna} and VATEX \cite{wang2019vatex}. We additionally evaluate HMDB51~\cite{hmdb51} as an independent all-layer classification replication and UCF101 zero-shot classification over complete text–video layer pairs for the five shared-space models. 
Classification uses eight-shot linear probing and accuracy; clustering uses MiniBatchKMeans and V-measure; retrieval ranks videos by cosine similarity and reports mean nDCG@10. Classification and clustering use ten evaluation seeds.

\paragraph{Geometry pools.} Classification and clustering geometry is computed from the corresponding training representations rather than the evaluation labels. Retrieval geometry uses the target query and corpus representations and is therefore target-specific, but does not access relevance judgments.

\subsection{Geometric measurements} 
\label{sec:geometric_measurements} 

\paragraph{Local geometry and Local Instrinsic Dimensionality.} For an embedding \(z\), let \(r_i(z)\) be its Euclidean distance to its \(i\)-th nearest neighbor (excluding itself). Each coordinate of the layer matrix is standardized to zero mean and unit population variance before nearest-neighbor search. We use the bias-corrected maximum-likelihood estimator \cite{NIPS2004_74934548,lid-amsaleg1028} 
\begin{equation} \widehat d(z) = \left[ \frac{1}{k-2} \sum_{i=1}^{k-1} \log\frac{r_k(z)}{r_i(z)} \right]^{-1}, \qquad k=100. \label{eq:lid} \end{equation} 

The layer value is the harmonic mean of the finite, strictly positive pointwise estimates. LID describes local neighborhood complexity and is distinct from representation quality. Sensitivity to \(k\) is reported in the \Cref{app:lid_k_sensitivity}. 

\paragraph{Global geometry and RankMe.} 
We compute RankMe~\cite{rankme-garrido-lecun-2023} from the singular values of the native representation matrix . We additionally compute covariance effective rank from the squared singular values after coordinate standardization and centering~\cite{roy2007effectiverank}. The former describes the native singular spectrum; the latter describes the standardized covariance spectrum. Implementation details and the normalization used for visualization are given in \Cref{app:spectral_measure_details}. 

\subsection{Cross-modal geometry} 
\label{sec:cross_modal_geometry} 

For a shared-space model \(m\), let \(B_m^{(T)}\) and \(B_m^{(V)}\) denote the numbers of transformer blocks in the text and video streams, respectively. Their normalized layer depths are \(\tau_{\ell_T}=\ell_T/B_m^{(T)}\) and \(\tau_{\ell_V}=\ell_V/B_m^{(V)}\). 
We evaluate every combination of text and video layers. To compare streams with different number of layers, we also match each layer in one stream to the layer in the other stream at the nearest normalized depths.

We compare text and video marginal distributions using empirical SW$_2$ distance. For equally sized matrices \(X,Y\in\mathbb{R}^{n\times D}\) and unit projection directions \(\theta_q\), 
 
 \begin{equation}
\operatorname{SW}_{2}(X,Y) =
\left[\frac{1}{M}
\sum_{q=1}^{M}\frac{1}{n}
\sum_{i=1}^{n}\left(
X^{(q)}_{(i)}-Y^{(q)}_{(i)}
\right)^2
\right]^{1/2},
\end{equation}

where \(X^{(q)}_{(i)}\) and \(Y^{(q)}_{(i)}\) are the ordered projections onto \(\theta_q\). Vectors receive the same row-wise normalization as the retrieval scorer and are not coordinate-standardized. We use \(M=1{,}024\) fixed Gaussian directions for each model-task grid. We measure paired structure separately using held-out orthogonal Procrustes error since SW$_2$ depends only on the marginal distributions, it is unchanged by reassigning the correspondence between text and video examples.

Given paired text and video representations, we use orthogonal Procrustes to fit a rotation between the two spaces on a training subset and then measure the alignment error on held-out pairs. Lower error indicates that corresponding text and video representations share recoverable geometric structure under such orthogonal transformation. Unlike SW$_2$, this measure depends on the specific pairing between text and video examples. Additional details on SW$_2$ and Procrustes metrics are provided in \Cref{app:sw2_measure_details} and \Cref{app:procrustes_measure_details}, respectively. For each evaluated configuration across model, task and depth, we generate 1,000 permutations by reassigning text-video correspondence while leaving both marginal distributions unchanged.
\section{Results}
\label{sec:results}

First, we examine how representation geometry changes across layers, and whether these changes in geometry can be used to identify useful representations. Next, we evaluate every layer on single-modal tasks,classification and clustering, and multi-modal tasks, to text--video correspondence and retrieval. Full results are reported in the Appendix.

\subsection{Layerwise geometry differs across models}
\label{sec:results_model_profiles}

\begin{figure}[htb]
    \centering
    \includegraphics[width=\linewidth]{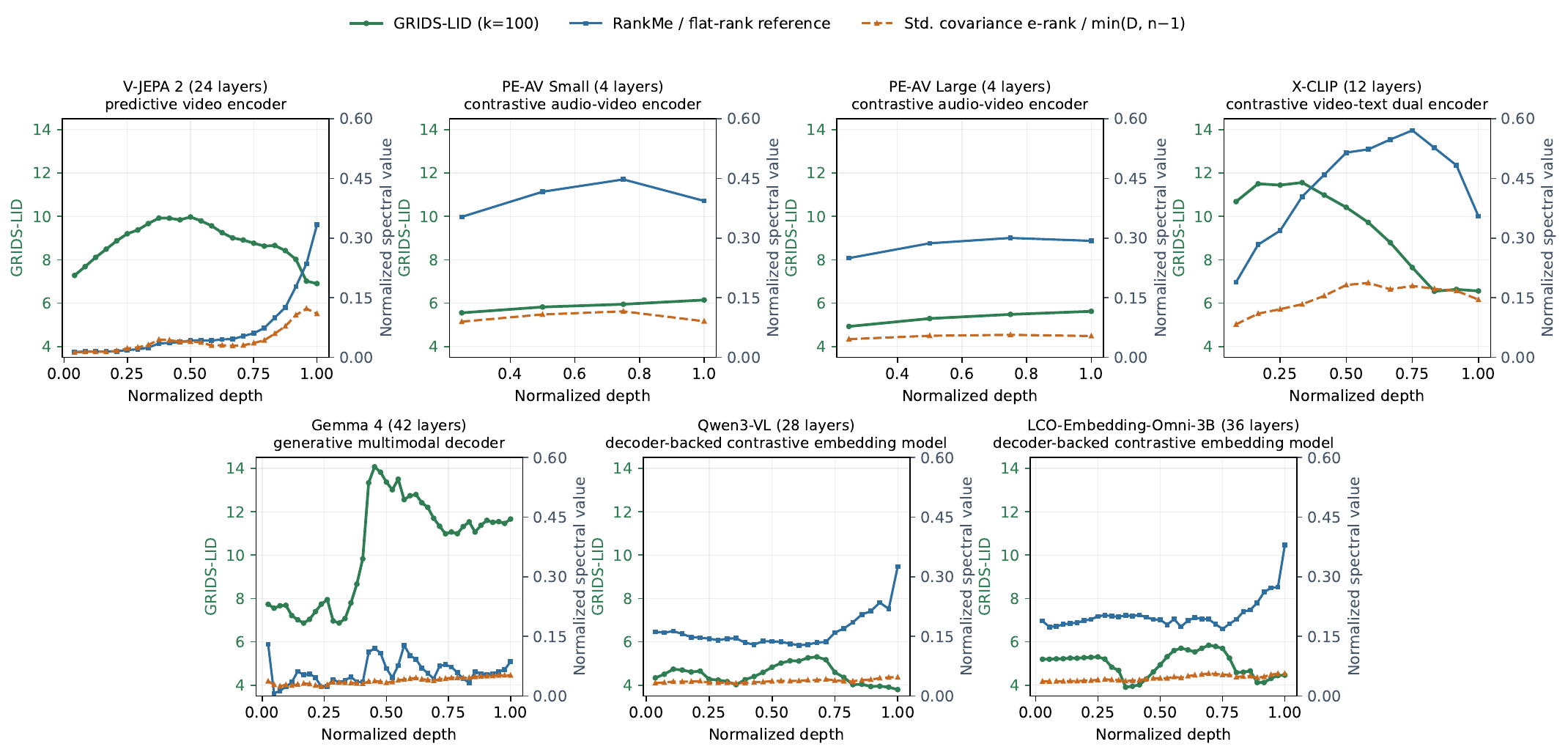}
    \caption{Full-pool UCF101 geometry signatures grouped by
    architecture and readout family. Spectral values are normalized for
    display only; accepted associations and selectors use their absolute
    values.}
    \label{fig:model_signatures}
\end{figure}

The models follow different geometric profiles across depth.
\Cref{fig:model_signatures} shows GRIDS-LID (\(k=100\)),
RankMe, and standardized covariance effective rank on UCF101. V-JEPA 2 has a broad mid-layer GRIDS-LID peak followed by a sharp decline, while RankMe rises near the end. X-CLIP shows a similar separation: GRIDS-LID peaks early, whereas RankMe peaks later. Both PE-AV models show increasing GRIDS-LID over their four video layers. Gemma 4 changes abruptly near the middle of the network, while Qwen3-VL and LCO show a late increase in RankMe as GRIDS-LID falls or remains non-monotonic.

Some similarities emerge within related model families: PE-AV Small and Large have similar GRIDS-LID profiles, and Qwen3-VL and LCO both show late RankMe growth. X-CLIP is also trained for cross-modal joint embedding, but follows a different trajectory for GRIDS-LID and RankMe. These are descriptive patterns rather than model-family effects as the models differ in architecture, data, scale, preprocessing, and output processing.

\subsection{No geometric measure identifies the best layer across all tasks}
\label{sec:results_geometry_performance}
No geometric measure provides a consistently reliable layer choice across all the evaluated model--task comparisons. Across the 21 Breakfast and UCF101 classification and clustering comparisons, choosing the layer with minimum GRIDS-LID gives a mean regret of 4.47\%, while choosing the layer with maximum GRIDS-LID gives 9.97\%. RankMe gives the lowest mean regret, 1.43\%, followed by the final-layer baseline at 2.49\% and standardized covariance effective rank at 3.99\%.

\begin{figure}[t]
    \centering
    \includegraphics[width=1\linewidth]{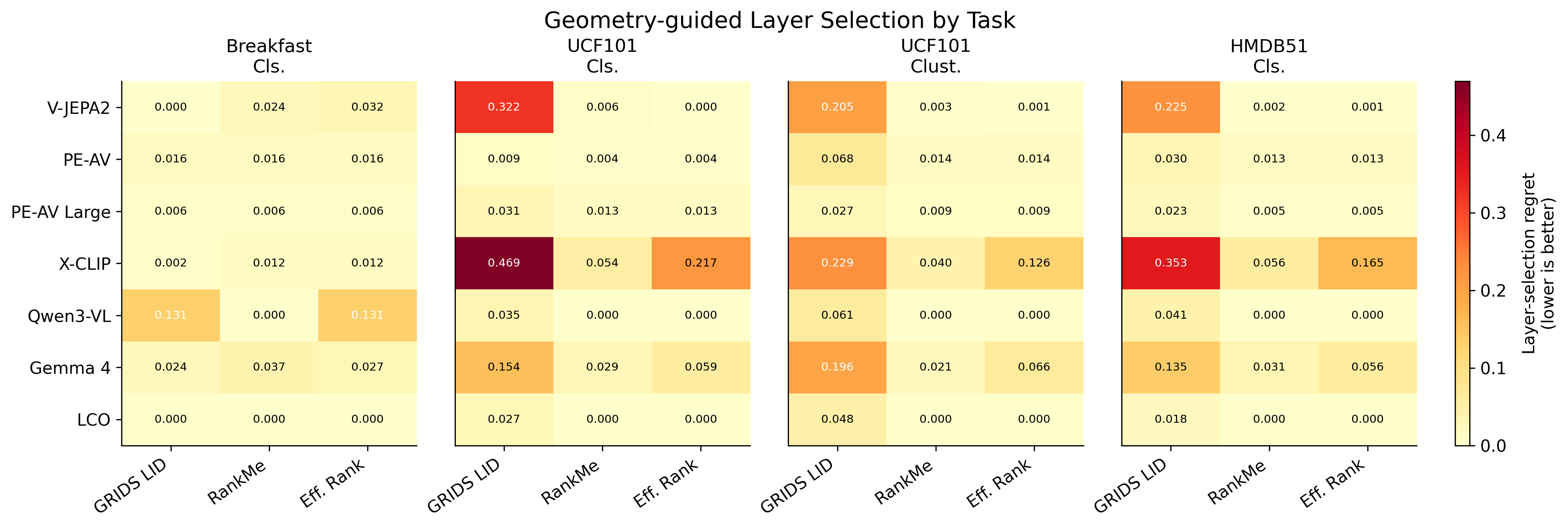}
    \caption{
        Layer-choice regret by task and model. Each cell shows the regret from selecting a layer using a geometric measure. Lower values are better; this makes it clear that the usefulness of a geometric measure depends on the task and model.}
    \label{fig:layervregret}
\end{figure}

\Cref{fig:layervregret} shows how these averages break down by task and model. GRIDS-LID varies in as a layer-selector on Breakfast and UCF101, with the favorable GRIDS-LID extremum changing across dataset and task. In several cases choosing the wrong extremum leads to large regret, specifically for V-JEPA 2 and X-CLIP on UCF101. The main pattern in GRID-LID's performance as a layer-selector is that it is dataset and task dependent, while RankMe is more stable across the same comparisons. Looking at the correlation between GRIDS-LID values for each model and task (\Cref{fig:app_geometry_associations}), we see positive correlations for some model–task pairs and negative correlations for others. For the LCO model, the Spearman correlation between GRIDS-LID and performance is $+0.942$ on Breakfast, $-0.375$ on UCF101 classification, and $-0.656$ on UCF101 clustering. GRIDS-LID-performance correlations range from $-0.935$ to $+0.962$ across model-task pairs. Across models and tasks, higher LID is not consistently associated with either better or worse downstream performance.

This task dependence is consistent with LID measuring local neighborhood complexity rather than task-specific properties. High LID may reflect useful fine-grained structure, but it may also arise from nuisance variation, mixed neighborhoods, or anomalies. Lower LID may similarly reflect useful abstraction or denoising, but can also arise from representation collapse. Therefore, neither higher nor lower LID is consistently associated with better downstream performance for the task in our study. Nevertheless, local dimensionality remains informative when neighborhood complexity or irregularity is itself the object of analysis, as in adversarial, backdoor, and anomaly-detection settings discussed in \Cref{related_work}.

We hypothesize RankMe performs better here because it measures global structure, which is more closely related to linear classification and distance-based clustering than a local neighborhood measure such as LID. This is visible in models such as V-JEPA 2 and X-CLIP, for which increases in RankMe values generally align closely with downstream improvement, unlike layerwise trajectories in GRIDS-LID. However, even in this case, RankMe does not necessarily provide a consistent best-performing layer of choice. This is shown in the regret scores out of 21 comparison: lower in 8, the same in 9, and higher in only 4.

\subsection{Intermediate layers often 
improve single-modality downstream performance}
\label{sec:results_layer_choice}

Because the geometric measures in \Cref{sec:results_geometry_performance} do not provide a universal measure for best layer selection, we next evaluate \textit{all} hidden state embeddings at every model layer directly. Evaluating every layer changes the preferred representation in most of the models we study. Across Breakfast classification, HMDB51 classification, and UCF101 classification and clustering, six of seven models have an intermediate layer that perform better than the final layer for at least one task. The largest gain occurs for Gemma 4 on UCF101 clustering, where layer 13 improves V-measure by  \(\Delta=+0.1683\) for UCF101 cluster and  \(\Delta=+0.0820 \) in HMDB51 classification over the last layer. PE-AV Large also improves at its first video layer on both UCF101 tasks by \(\Delta=+0.0315\) in classification accuracy and \(\Delta=+0.0268\) in V-measure. PE-AV Small improves performance over the last layer with a \(\Delta=+0.0297\) in the HMDB51 classification task. See Section \Cref{app:hmdb51_results} for more details.

\begin{figure}
    \centering
    \includegraphics[width=1\linewidth]{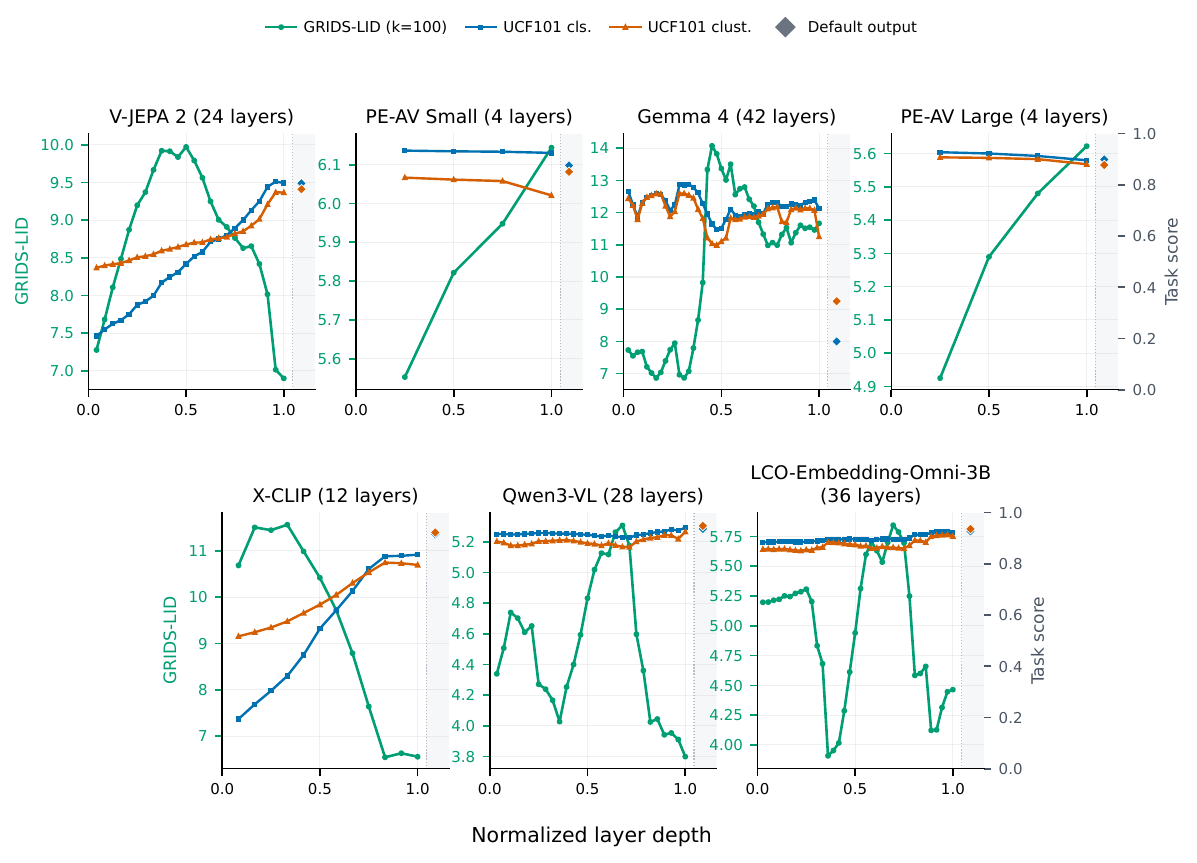}
    \caption{
    UCF101 layerwise profiles for seven models. GRIDS-LID, classification and clustering means are plotted layerwise against normalized model depth. For each model profile, diamonds in each shaded margin mark the model-default output. Task scores are on a 0–1 scale.
}
    \label{fig:ucf101_depth_profiles}
\end{figure}

\Cref{fig:ucf101_depth_profiles} shows that these gains on task performance do not follow one depth pattern. V-JEPA 2 \cite{assran2025vjepa2} and X-CLIP continue to improve task performance after GRIDS-LID has reached its maximum and begun to fall. The PE-AV models show a different pattern, where their best UCF101 performance occurs at the first video layer while GRIDS-LID increases with depth. Qwen3-VL performs best at its final layer, while LCO peaks only slightly earlier, at layers 34 and 35. Gemma 4 reaches its best UCF101 scores before the sharp increase in GRIDS-LID at the middle of the network.

These varying patterns across models opposes the idea of a universal intermediate-layer optimum or a universal GRIDS-LID direction. A layer’s task performance can improve as GRIDS-LID rises, falls, or remains nearly the same. The best layer depth for task depends on what the model is trained to produce. Gemma’s decoder layers support generation, while PE-AV is trained using scaled contrastive learning to align video, audio, and text. Neither objective directly optimizes a mean-pooled video representation for linear classification or clustering. Therefore, the best representation for the tasks may emerge earlier in the network.
Qwen3-VL and LCO, both trained as embedding models, reach their best layerwise performance at or near the end of the network.

We find that layer representations and model-default outputs also differ in performance. Across the 21 model-task comparisons, the best layer representations exceeds the model-default output in 100 cases, including nine intermediate layer gains across V-JEPA 2, PE-AV Small, PE-AV Large and Gemma 4. In contrast, the model-default output outperforms every evaluated layer representation for all three X-CLIP and LCO comparisons.
These results show that downstream performance depends on both network depth and model-default output; the preferred representation varies by model and task.

\subsection{Distributional distance tracks retrieval but does not determine it}
\label{sec:results_crossmodal}

Sliced Wasserstein (SW$_2$) distance provides a useful, but incomplete view of text-video retrieval. Using SW$_2$ to measure cross-modal distributional distance, we find a strong relationship between SW$_2$ and retrieval performance across layer pairs. Across 5 models and 2 retrieval benchmarks (TUNA and VATEX), SW$_2$ is negatively associated with nDCG@10 in 8 of the 10 evaluations, with median Spearman correlations of $-0.6924$ on TUNA and $-0.7700$ on VATEX. This negative correlation suggests that lower distributional distance is generally associated with higher retrieval performance. However, minimum SW$_2$ is not the best criterion for selecting best video-text layer for retrieval. Compared to the oracle layer pair, the models’ default outputs incur a mean regret of 0.0097, where as minimum-SW$_2$ layer pair output have a mean regret 0.2994,  as presented in \Cref{fig:retrieval_sw2}. 

These results suggest that retrieval depends on more than cross modal distributional distance. SW$_2$ measures the distance between the marginal text and video distributions while ignoring the correspondence between text-video pairs. Reassigning which text is associated with which video leaves SW$_2$ unchanged while disrupting text-video correspondence needed for retrieval.  To isolate this effect, we use paired permutations as a correspondence control. The cross-modal pairs  are randomly reassigned while the text and video embeddings are unchanged. Held-out orthogonal Procrustes error and retrieval performance are evaluated to identify if both metrics favor the true pairings over the randomized pairings. 
 
The correspondence control and orthogonal Procrustes error show that the original text-video pairs contain shared structure.  Across the evaluated models, layer depths, and tasks, held-out orthogonal Procrustes error favors the true pairing in 31 of 32 comparisons. Retrieval is a more stringent evaluation because cross-modal structure must translate to correct ranking . The true pairing has higher nDCG@10 than the randomized pairing ($p \leq 0.05$) in 27 of the 32 comparisons, with failures concentrated at earlier PE-AV Large and X-CLIP layer depth. \Cref{fig:app_pairing_controls} shows that at the final evaluated depth, every correspondence-sensitive metric and retrieval score distinguishes the true pairing from all 1,000 permutations. These results show that distributional distance, paired cross-modal structure, and ranked retrieval performance capture complementary characteristics of cross-modal representations (\Cref{fig:retrieval_sw2}). 

\subsection{Cross-modal retrieval peaks late, but the best text and video depths may differ}
\label{sec:results_retrieval_depth}

\begin{figure}
    \centering
    \includegraphics[width=01\linewidth]{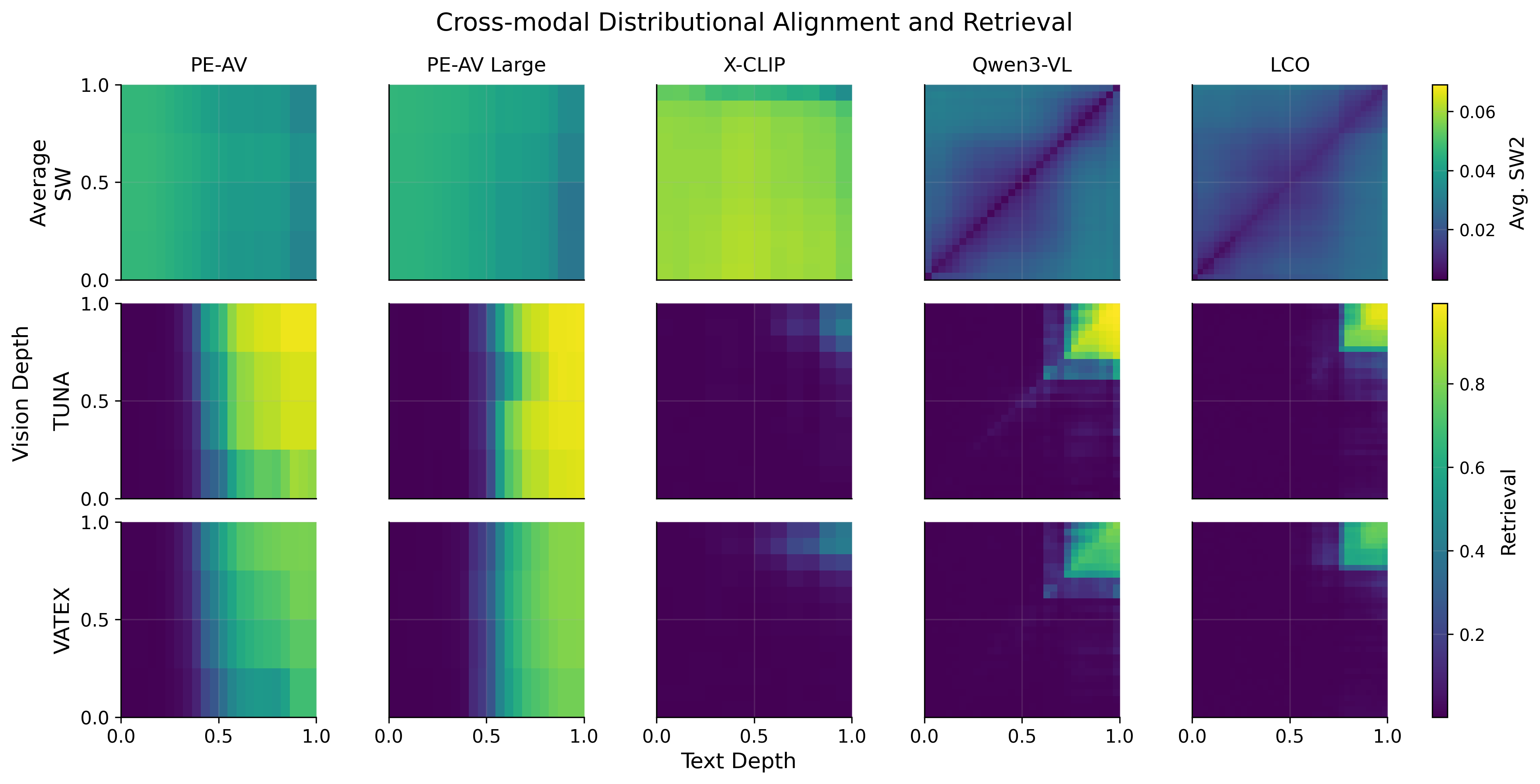}
    \caption{Cross-modal distributional alignment and retrieval performance across normalized text and vision depth. The top row shows average Sliced Wasserstein distance (SW$_2$) between text and vision representations, while the bottom rows show text-to-video retrieval performance (nDCG@10) on TUNA and VATEX. Although lower SW$_2$ often coincides with improved retrieval, we find that cross-modal layer pairs, with minimum SW$_2$, do not consistently result in a high retrieval performance.}
    \label{fig:retrieval_sw2}
\end{figure}

For each model and retrieval task, we define the best layer pair as the raw text–video layer pair with the highest nDCG@10. Unlike modality specific task layer performance as shown in \Cref{sec:results_layer_choice}, the cross-modal retrieval task prefers final and near-final layer pairs. One possible explanation is that retrieval depends more strongly on the model’s final processing steps that turn hidden states into retrieval embeddings, but we do not isolate this mechanism here.

In most evaluations, the best pair also lies on the matched-depth diagonal. The main exceptions are X-CLIP and LCO. For X-CLIP, using the final text layer and penultimate video layer improves performance over the best matched-depth pair ($\Delta = +0.0685$ nDCG@10 on TUNA; $\Delta = +0.0198$ on VATEX), while LCO reaches its best VATEX score at text layer 34 and video layer 35 ($\Delta = +0.0062$). These exceptions suggest that the most retrieval performance text and video layers do not always occur at matched depths. Full layer-pair results are reported in \Cref{tab:app_retrieval_layer_pairs}.

The SW$_2$ heatmaps in~\Cref{fig:retrieval_sw2}  show that models with similar retrieval performance can have different cross-modal geometries. PE-AV and X-CLIP show low distributional distance between text and video embeddings across a broad range of layer pairs. On the other hand, Qwen3-VL and LCO reach their lowest SW$_2$ distances at matched text–video layers. Measuring layer-wise SW$_2$ distance reveals differences in cross-modal organization throughout the model that are not visible from evaluating retrieval performance alone. Because the models vary in architecture, training setup, and output processing, we cannot attribute these patterns solely to their training objectives.
\section{Discussion and Conclusion}
We introduce \layerscope, a layerwise diagnostic framework for comparing representation geometry, downstream performance, and layer-selection across models and network depth. Layerwise evaluation shows that intermediate layers often improve single modality task performance for classification and clustering, while cross-modal retrieval often performs well at final or near-final text-video layer pairs.  At a descriptive level, layerwise geometric trajectories differ substantially across model families and thus provide a label-free approach to characterize representation changes and behavior throughout the network. When using geometric metrics for layer selection, the layer of choice depends on the downstream task and model.  In our results of both local and global measurements,
GRIDS-LID exhibits task-dependent relationships with downstream performance, so neither higher nor lower local dimensionality values associate with a best-performing layer for all tasks. 

We have empirically shown that across a subset of MVEB tasks, RankMe has the lowest mean layer-selection regret among the geometric measures tested for classification and clustering comparisons; however, it does not generalize as a universal layer-selector. This shows that association with downstream performance does not necessarily imply reliable layer selection, and that the preferred representation depends on the model and task. Geometry is therefore most useful as a characterization tool to help identify where representations change and provides descriptive numerical measures for these changes, but it does not replace task-specific layer evaluation.
In our retrieval analysis, we have first shown that a lower distribution distance, measured by SW$_2$, is associated with higher nDCG@10, but does not generally identify the best text-video layer pair. For 1,000 shuffle pairings, Procrustes error has identified 31 of 32 text-video comparisons across models, datasets, and network depth, revealing paired structure that SW$_2$ cannot support.
Future work should use controlled comparisons to isolate how design choices such as training objective, architecture, data scale, and output mapping shape these layerwise geometric profiles.

\clearpage
\bibliographystyle{plainnat}
\bibliography{references}

\clearpage
\appendix
\crefalias{section}{appendix}
\crefname{appendix}{appendix}{appendices}
\Crefname{appendix}{Appendix}{Appendices}
\section{Experimental Details and Reproducibility}
\label[appendix]{app:details}

\Crefrange{tab:model_inventory}{tab:task_scope} summarize the models,evaluated layers, and downstream tasks. Definitions and implementation details for the geometric measures are given in
\Cref{app:6b_geometric_measure_details}.

\begin{table}[H]
    \centering
    \scriptsize
    \setlength{\tabcolsep}{3pt}
    \caption{Models and checkpoints used in the experiments. Modalities indicate the inputs used in this study.}
    \label{tab:model_inventory}
    \resizebox{\textwidth}{!}{%
    \begin{tabular}{@{}lllll@{}}
        \toprule
        Model Family & Model & Checkpoint & Modalities & Architecture \\
        \midrule
        Predictive video & V-JEPA 2 & \texttt{facebook/vjepa2-vitl-fpc64-256} & Video & ViT-L \\
        Audio-video contrastive & PE-AV Small & \texttt{facebook/pe-av-small} & Text/video & Video + text encoders \\
        Audio-video contrastive & PE-AV Large & \texttt{facebook/pe-av-large} & Text/video & Video + text encoders \\
        Video-text contrastive & X-CLIP & \texttt{microsoft/xclip-base-patch32} & Text/video & Base dual encoder \\
        Autoregressive multimodal & Gemma 4 & \texttt{google/gemma-4-E4B-it} & Multimodal & E4B-IT decoder \\
        Autoregressive multimodal & Qwen3-VL & \texttt{Qwen/Qwen3-VL-Embedding-2B} & Text/video & Embedding-2B decoder \\
        Autoregressive multimodal & LCO & \texttt{LCO-Embedding/LCO-Embedding-Omni-3B} & Omnimodal & Embedding-Omni-3B decoder \\
        \bottomrule
    \end{tabular}
    }
\end{table}

\begin{table}[H]
    \centering
    \small
    \setlength{\tabcolsep}{5pt}
    \caption{Evaluated layer ranges and retrieval eligibility.}
    \label{tab:layer_coverage}
    \begin{adjustbox}{max width=\textwidth}
    \begin{tabular}{@{}lllll@{}}
        \toprule
        Model & \# layers scored & Scored layers & Retrieval eligible? & Notes \\
        \midrule
        V-JEPA 2 & 24 video & Video 1--24 & No & Classification/clustering \\
                Gemma 4 & 42 video & Video 1--42 & No & Classification/clustering \\
        PE-AV Small & 4 video / 22 text & Video 1--4; text 1--22 & Yes & Retrieval \\
        PE-AV Large & 4 video / 22 text & Video 1--4; text 1--22 & Yes & Retrieval \\
        X-CLIP & 12 video / 12 text & Video 1--12; text 1--12 & Yes & Retrieval \\
        Qwen3-VL & 28 video / 28 text & Video 1--28; text 1--28 & Yes & Retrieval \\
        LCO & 36 video / 36 text & Video 1--36; text 1--36 & Yes & Retrieval \\
        \bottomrule
    \end{tabular}
    \end{adjustbox}
\end{table}

We keep all model parameters frozen throughout the experiments. The model-default output is evaluated separately from the final-layer representation as it may have additional normalization, pooling, projection etc.

\paragraph{Representation and task scope.}
For classification and clustering, we evaluate every listed video layer, constructing its representation by averaging the eligible hidden states before the model-specific output mapping, excluding padding and special tokens. For LCO, this average includes only eligible modality tokens.

For retrieval and zero-shot classification, we evaluate every text--video layer pair for PE-AV Small, PE-AV Large, X-CLIP, Qwen3-VL, and LCO. At each layer, we apply the normalization, pooling, and projection operations of the model's frozen shared embedding interface. We then compare the resulting text and video vectors using cosine similarity.
No layer-specific projections or cross-modal alignment adapters are trained.

\begin{table}[htb]
    \centering
    \caption{
        Downstream evaluations. Classification and clustering scores are summarized over ten evaluator seeds. Retrieval and zero-shot classification are deterministic for fixed representations.
    }
    \label{tab:task_scope}
    \scriptsize
    \setlength{\tabcolsep}{3.4pt}
    \renewcommand{\arraystretch}{1.10}
    \resizebox{\linewidth}{!}{
    \begin{tabular}{lllll}
        \toprule
        Analysis &
        Dataset &
        Evaluated representations &
        Evaluation &
        Score \\
        \midrule
        Classification &
        Breakfast &
        Every video layer; model-default output &
        Eight-shot linear probe &
        Accuracy \\
        Classification &
        UCF101 &
        Every video layer; model-default output &
        Eight-shot linear probe &
        Accuracy \\
        Classification &
        HMDB51 &
        Every video layer; model-default output &
        Eight-shot linear probe &
        Accuracy \\
        Clustering &
        UCF101 &
        Every video layer; model-default output &
        MiniBatchKMeans &
        V-measure \\
        Text-to-video retrieval &
        TUNA-Bench &
        Every text--video layer pair; model-default pair &
        Cosine ranking &
        Mean nDCG@10 \\
        Text-to-video retrieval &
        VATEX &
        Every text--video layer pair; model-default pair &
        Cosine ranking &
        Mean nDCG@10 \\
        Zero-shot classification &
        UCF101 &
        Every text--video layer pair; model-default pair &
        Fixed four-prompt ensemble &
        Accuracy \\
        \bottomrule
    \end{tabular}}
\end{table}

We compute classification and clustering geometry from the corresponding training representations without using class labels. Retrieval geometry is specific to the target dataset: it uses the query and video representations but not relevance judgments.

\section{Geometric measure definitions and implementation details}
\label{app:6b_geometric_measure_details}

Let \(X_\ell\in\mathbb{R}^{n\times D}\) denote the representation matrix at layer \(\ell\), with one row per example. For classification and clustering, each row is the pooled layer representation defined in \Cref{sec:models_representations_tasks}. For retrieval, each row is the representation produced by the model-specific shared-space readout at that layer. Unless stated otherwise, each geometric measure is computed separately at every evaluated layer using the full representation pool. Each metric applies its own preprocessing to \(X_\ell\). 
GRIDS-LID standardizes each feature coordinate before nearest-neighbor search. RankMe is computed directly from \(X_\ell\), without additional centering or coordinate standardization. Standardized covariance effective rank first standardizes and centers \(X_\ell\). Cross-modal measures use the row-wise normalization applied by the retrieval interface. 

\subsection{GRIDS-LID, depth roughness and neighborhood-size sensitivity}
\label{app:lid_k_sensitivity}

For a representation $z \in X_\ell$, let \(r_i(z)\) denote the Euclidean distance from $z$ to its $i$-th nearest neighbor, excluding $z$ itself. Before nearest-neighbor search, each feature coordinate of \(X_\ell\) is standardized across the $n$ examples to zero mean and unit population variance. We use the bias-corrected maximum-likelihood estimator \cite{lid-amsaleg1028,NIPS2004_74934548}

\begin{equation}
\widehat d(z)
=
\left[
\frac{1}{k-1}
\sum_{i=1}^{k-1}
\log\frac{r_k(z)}{r_i(z)}
\right]^{-1},
\qquad k=100.
\label{eq:app_lid_pointwise}
\end{equation}

The $k$-th neighbor defines the radius of the neighborhood, and the estimator uses
the first $k-2$ neighbor distances within that radius.
\footnote{\textit{Note }that GRIDS-LID uses $k-1$ neighbors. However, we use $k-2$ with only drops that are \textit{less than 1\%}, thus rendering the choice statistically negligible \cite{rozza2012_dim_estimators}.}

Let \(\mathcal{S}_{\ell}^{+}\) be the finite, strictly positive pointwise estimates at layer \(\ell\). The layer-level GRIDS-LID value is their harmonic mean,
\begin{equation}
d_\ell
=
\left[
\frac{1}{|\mathcal{S}_{\ell}^{+}|}
\sum_{z\in\mathcal{S}_{\ell}^{+}}
\frac{1}{\widehat d(z)}
\right]^{-1}.
\label{eq:app_lid_layer}
\end{equation}

Nonfinite and nonpositive estimates are excluded from the layer-level aggregation. We
evaluate both the minimum and maximum GRIDS-LID layers without assuming either direction as generally preferable.

\paragraph{LID depth roughness.}
We also measure local variation in GRIDS-LID across network depth. Let
$W_\ell$ denote the contiguous odd-width window of neighboring layers
associated with layer $\ell$, and let

\begin{equation}
m_\ell = \operatorname{median}_{j\in W_\ell} d_j .
\label{eq:median_depth}
\end{equation}

LID depth roughness is
\begin{equation}
\rho_{\mathrm{LID}}(\ell)
=
\frac{
\operatorname{median}_{j\in W_\ell}
|d_j-m_\ell|
}{
|m_\ell| + 10^{-12}
}.
\label{eq:lid_depth_roughness}
\end{equation}

Lower roughness values indicate less variation in GRIDS-LID over neighboring layers. The primary analysis uses windows spanning approximately $20\%$ of model depth -- $10\%$ and $30\%$ windows are used for sensitivity analysis.

Performance depth roughness is computed analogously from the mean downstream score at each layer. It is only used to compare local variation in geometry with local variation in downstream performance. We only analyze roughness-based performance for models with at least ten evaluated layers. 

\paragraph{Neighborhood-size sensitivity.}
We repeat the GRIDS-LID calculation at $k\in\{50,150\}$ using the same representation matrices, coordinate standardization, Euclidean distance, and layer-level aggregation as in the primary $k=100$ analysis. 

\begin{figure}[htb]
    \centering
    \includegraphics[width=0.88\linewidth]
    {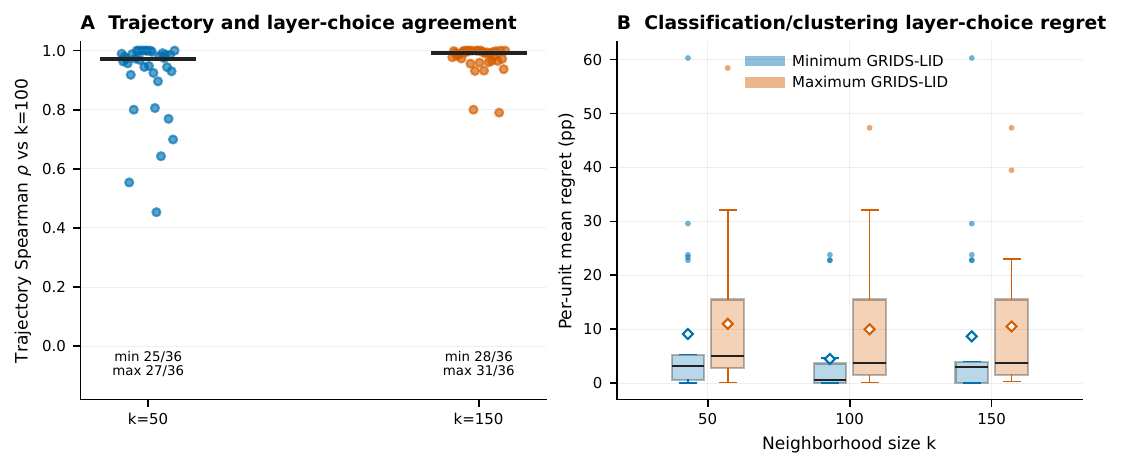}
    \caption{
        Sensitivity of GRIDS-LID to neighborhood size. Panel A compares the \(k=50\) and \(k=150\) trajectories with the primary \(k=100\) trajectory over 36 representation views and reports agreement of the minimum- and maximum-GRIDS-LID layers.
        Panel B reports layer-choice regret over the 21 primary
        classification and clustering comparisons, using evaluator seeds $0$--$9$. Regret is shown on the $0$--$1$ downstream-score scale.
    }
    \label{fig:app_lid_k_sensitivity}
\end{figure}

\subsection{Spectral measures}
\label{app:spectral_measure_details}

\paragraph{RankMe.}
RankMe is computed from the singular values of the native representation matrix \(X_\ell\), without additional centering or coordinate standardization \cite{rankme-garrido-lecun-2023}.

Following the released implementation, 

\begin{equation}
p_i
=
\frac{\sigma_i(X_\ell)}
     {\sum_j \sigma_j(X_\ell)},
\qquad
\widetilde p_i
=
p_i+10^{-7}.
\label{eq:app_rankme_weights}
\end{equation}

and

\begin{equation}
\operatorname{RankMe}(X_\ell)
=
\exp\left(
-\sum_i
\widetilde p_i\log\widetilde p_i
\right).
\label{eq:app_rankme}
\end{equation}

The numerical offset is added after normalizing the singular values, and the resulting values are not renormalized. 

\paragraph{Standardized covariance effective rank.}
Covariance effective rank is computed after coordinate standardization and centering. If \(s_i\) are the singular values of the resulting matrix, let
\begin{equation}
q_i
=
\frac{s_i^2}{\sum_j s_j^2}.
\label{eq:app_covrank_weights}
\end{equation}

The effective rank \cite{roy2007effectiverank} is 
\begin{equation}
r_{\mathrm{cov}}(X_\ell)
=
\exp\left(
-\sum_{i:q_i>0}
q_i\log q_i
\right).
\label{eq:app_covrank}
\end{equation}

The analysis uses the absolute RankMe and covariance-effective-rank values. For display only, RankMe is divided by the value obtained from a flat singular-value spectrum of the same
length, and covariance effective rank is divided by $\min(D,n-1)$. Thus, RankMe retains the feature scaling present in $X_\ell$, whereas standardized covariance effective rank is computed from the variance spectrum after coordinate standardization and centering.

\subsection{Sliced Wasserstein distance and relative-depth alignment}
\label{app:sw2_measure_details}

For a shared-space model, let \(X,Y\in\mathbb{R}^{n\times D}\) contain text and video representations from the same retrieval evaluation pool. We $\ell_2$-normalize each row using the same normalization as the retrieval scorer, while feature coordinates are not standardized.

For unit directions \(\theta_1,\ldots,\theta_M\), empirical sliced \(2\)-Wasserstein distance \cite{nietert2022slicedwasserstein} is 
\begin{equation}
\operatorname{SW}_{2}(X,Y)
=
\left[
\frac{1}{M}
\sum_{q=1}^{M}
\frac{1}{n}
\sum_{i=1}^{n}
\left(
X^{(q)}_{(i)}
-
Y^{(q)}_{(i)}
\right)^2
\right]^{1/2},
\label{eq:app_sw2}
\end{equation}

where \(X^{(q)}_{(i)}\) and \(Y^{(q)}_{(i)}\) are the \(i\)-th ordered one-dimensional projections of the two sets of representations onto \(\theta_q\). 

For each model--task comparison, we use $M=1024$ fixed Gaussian directions normalized to unit length, and re-use the same directions for every evaluated text--video layer pair. SW$_2$ compares the empirical distributions of text and video representations without using which text corresponds to which video. Therefore, and importantly, permutation of the text--video correspondence leaves SW$_2$ unchanged.

When comparing text and video layers at corresponding relative depths, text
layer $\ell_T$ is paired with the video layer whose relative position in the
network is closest
\[
\ell_V^\dagger(\ell_T)
=
\arg\min_{\ell_V}
\left|
\frac{\ell_T}{B_m^{(T)}}
-
\frac{\ell_V}{B_m^{(V)}}
\right|.
\]

\subsection{Correspondence-sensitive measures}
\label{app:correspondence_measure_details}

Let
\[
X=[x_1,\ldots,x_n]^\top,
\qquad
Y=[y_1,\ldots,y_n]^\top
\]

contain paired and row-normalized text and video representations, where \(x_i\) and \(y_i\) describe the same example. 
Unlike SW\(_2\), the measures in this section depend on this paired example correspondence and therefore, change when we permute this text–video pairing.

\paragraph{Held-out orthogonal Procrustes error.}
\label{app:procrustes_measure_details}
We fit and evaluate the paired examples with an 80\%/20\% split respectively. We then estimate the modality means \(\mu_X\) and \(\mu_Y\) from the fit set and used to center both the fit and evaluation representations,
\[
X_{\mathrm{fit}}^c
=
X_{\mathrm{fit}}-\mathbf{1}\mu_X^\top,
\qquad
Y_{\mathrm{fit}}^c
=
Y_{\mathrm{fit}}-\mathbf{1}\mu_Y^\top.
\]

Finally, we fit an unscaled orthogonal map
\begin{equation}
R^\star
=
\arg\min_{R^\top R=I}
\left\|
X_{\mathrm{fit}}^cR-Y_{\mathrm{fit}}^c
\right\|_F^2.
\label{eq:app_procrustes_fit}
\end{equation}
The held-out error is the root-mean-square Euclidean residual,
\begin{equation}
E_{\mathrm{Proc}}
=
\frac{
\left\|
(X_{\mathrm{eval}}-\mathbf{1}\mu_X^\top)R^\star
-
(Y_{\mathrm{eval}}-\mathbf{1}\mu_Y^\top)
\right\|_F
}{
\sqrt{n_{\mathrm{eval}}}
}.
\label{eq:app_procrustes_error}
\end{equation}
Lower values indicate that a single orthogonal alignment fitted on one
set of text--video pairs generalizes more accurately to unseen pairs.

\subsection{Correspondence permutations}
\label{app:correspondence_permutations}

The registered correspondence analysis covers the $32$ model--task--depth comparisons, more details in \Cref{app:paired_correspondence_details}. For each comparison, we construct $1000$ permutations by shuffling the correspondence between text and video examples while leaving both sets of representations unchanged. The permutation therefore changes item pairing without changing either marginal empirical distribution.

For measure \(g\), the one-sided Monte Carlo value is
\begin{equation}
p_{\mathrm{MC}}
=
\frac{1+b}{1001},
\label{eq:app_mc}
\end{equation}

where \(b\) is the number of permuted values at least as favorable as the observed value. Larger values are favorable for cosine margin, linear CKA, neighbor overlap, and nDCG@10; smaller values are favorable for Procrustes error. Note that the minimum attainable value is \(1/1001\), when the observation is more favorable than all 1,000 permutations.

\Cref{fig:app_pairing_controls} illustrates the separation between the observed text-video pairing from the permutations. For a measure with permutation mean \(\mu_{\mathrm{perm}}\) and standard deviation
\(\sigma_{\mathrm{perm}}>0\), we define
\begin{equation}
z_{\mathrm{dir}}
=
s_g
\frac{
g_{\mathrm{obs}}-\mu_{\mathrm{perm}}
}{
\sigma_{\mathrm{perm}}
},
\label{eq:app_directional_z}
\end{equation}
where \(s_g=+1\) for larger-is-better measures and \(s_g=-1\) for
Procrustes error. A positive value means the original text–video pairing gives a better result than the shuffled pairings on average.

\section{Cross-modal task details}
\label{app:crossmodal_task_results}

\subsection{Sliced Wasserstein distance and retrieval}
\label{app:sw2_retrieval_details}

For each shared-space model, we evaluate all text--video layer pairs $p=(\ell_T,\ell_V)$ on TUNA and VATEX. The model-default text and video outputs are evaluated separately. We also compare text and video layers at corresponding relative depths, pairing each text layer with the video layer at the closest relative position in the network.

\begin{figure}[t]
    \centering
    \includegraphics[width=\linewidth]
    {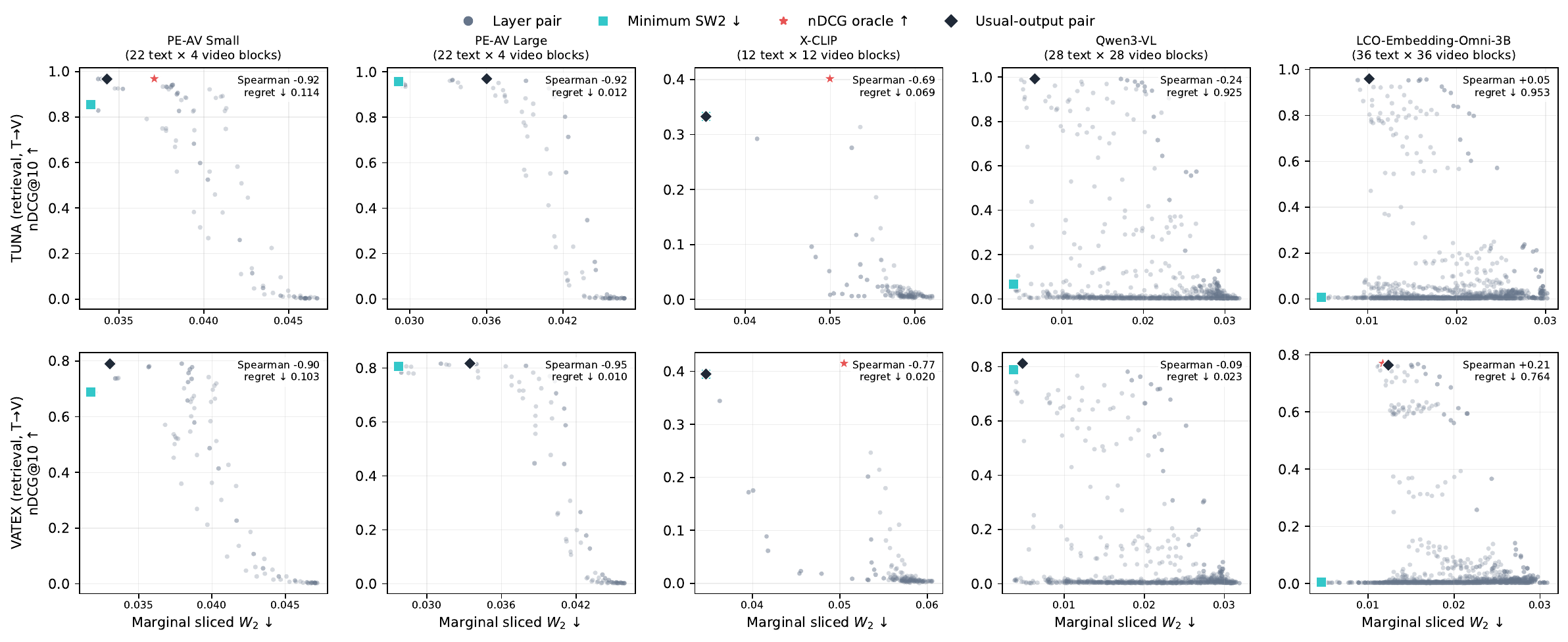}
    \caption{
        $SW_2$ and nDCG@10 across all $5218$ accepted layer-pair cells. Cyan squares mark minimum SW$_2$, red stars the retrieval oracle, and black diamonds the detached default pair. Lower marginal distance often tracks retrieval but does not identify the optimum; axes are model-specific.
    }
    \label{fig:app_sw2_ndcg_scatter}
\end{figure}

\Cref{fig:app_sw2_ndcg_scatter} compares SW$_2$ and nDCG@10 across all evaluated text-video layer-pairs.
SW$_2$ is negatively associated with nDGC@10 in eight of the ten model-task comparisons.
However the layer pair with the minimum SW$_2$ is never the best performing retrieval pair, its mean selection regret is $0.2994$, compared with $0.0097$ for the model=default pair. The mean Spearman correlation across the ten comparisons is $-0.514$, decreasing to $-0.279$ after accounting for relative text and video depth.

\Cref{tab:app_retrieval_layer_pairs} reports the best layer pair, the best matched-depth score, and the model-default score. The differences are
computed relative to the retrieval oracle before the displayed values are rounded.

\begin{table}[t]
    \centering
    \caption{
        Text-to-video retrieval over all text--video layer pairs. The
        oracle is the pair with the highest nDCG@10. The matched-depth
        optimum is selected from the normalized-depth path, while the
        model-default output is evaluated separately. Scores and differences
        use the \(0\)--\(1\) nDCG@10 scale.
    }
    \label{tab:app_retrieval_layer_pairs}
    \scriptsize
    \setlength{\tabcolsep}{3.1pt}
    \renewcommand{\arraystretch}{1.08}
    \resizebox{\linewidth}{!}{
    \begin{tabular}{llccccccc}
        \toprule
        Task &
        Model &
        Matrix &
        Oracle \((\ell_T,\ell_V)\) &
        Oracle score &
        Matched depth &
        \(\Delta_{\mathrm{matched}}\) &
        Model-default &
        \(\Delta_{\mathrm{default}}\) \\
        \midrule
        TUNA &
        PE-AV Small &
        \(22\times4\) &
        \(19/4\) &
        0.9681 &
        0.9681 &
        +0.0000 &
        0.9671 &
        +0.0010 \\
        VATEX &
        PE-AV Small &
        \(22\times4\) &
        \(21/4\) &
        0.7914 &
        0.7914 &
        +0.0000 &
        0.7899 &
        +0.0016 \\
        TUNA &
        PE-AV Large &
        \(22\times4\) &
        \(22/4\) &
        0.9695 &
        0.9695 &
        +0.0000 &
        0.9695 &
        +0.0000 \\
        VATEX &
        PE-AV Large &
        \(22\times4\) &
        \(22/4\) &
        0.8172 &
        0.8172 &
        +0.0000 &
        0.8172 &
        +0.0000 \\
        TUNA &
        X-CLIP &
        \(12\times12\) &
        \(12/11\) &
        0.4014 &
        0.3329 &
        +0.0685 &
        0.3328 &
        +0.0686 \\
        VATEX &
        X-CLIP &
        \(12\times12\) &
        \(12/11\) &
        0.4148 &
        0.3950 &
        +0.0198 &
        0.3950 &
        +0.0198 \\
        TUNA &
        Qwen3-VL &
        \(28\times28\) &
        \(28/28\) &
        0.9924 &
        0.9924 &
        +0.0000 &
        0.9924 &
        +0.0000 \\
        VATEX &
        Qwen3-VL &
        \(28\times28\) &
        \(28/28\) &
        0.8124 &
        0.8124 &
        +0.0000 &
        0.8124 &
        +0.0000 \\
        TUNA &
        LCO &
        \(36\times36\) &
        \(36/36\) &
        0.9592 &
        0.9592 &
        +0.0000 &
        0.9592 &
        +0.0000 \\
        VATEX &
        LCO &
        \(36\times36\) &
        \(34/35\) &
        0.7698 &
        0.7636 &
        +0.0062 &
        0.7636 &
        +0.0062 \\
        \bottomrule
    \end{tabular}}
\end{table}

\subsection{Cross-modal correspondence across depth}
\label{app:paired_correspondence_details}

\begin{figure}[t]
    \centering
    \includegraphics[width=\linewidth]
    {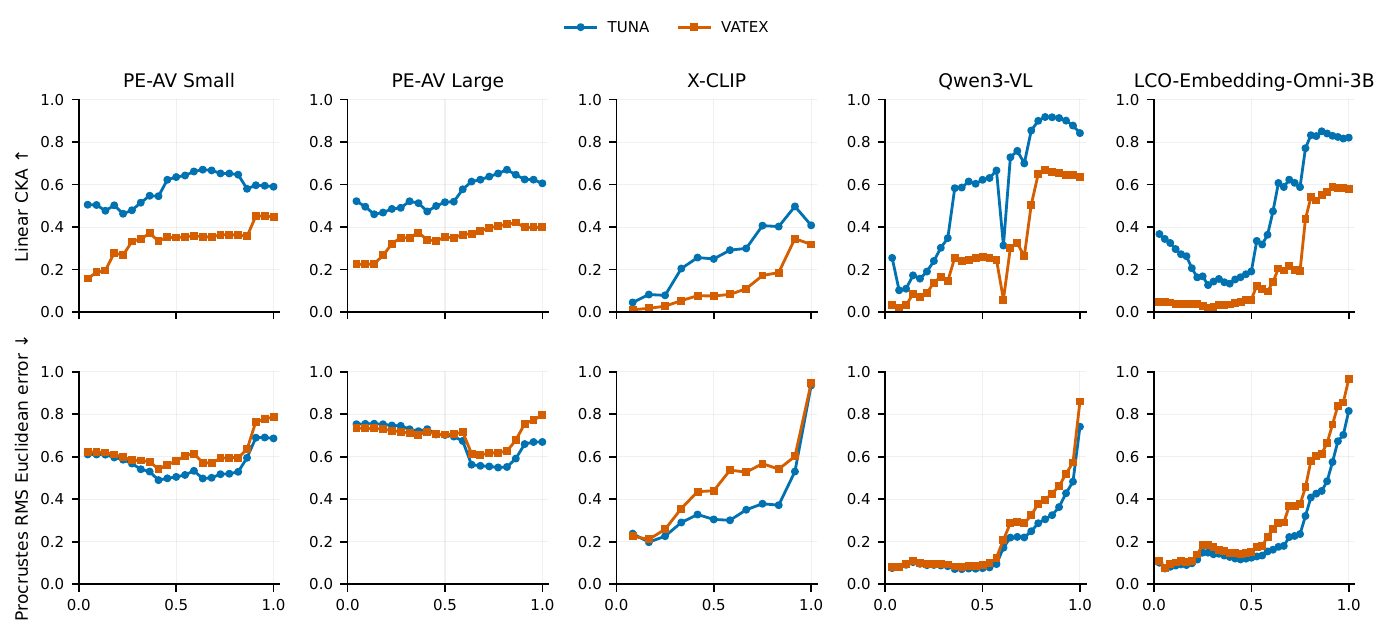}
    \caption{
        Linear CKA and held-out orthogonal Procrustes error for text–video layer pairs aligned by relative depth on TUNA and VATEX. Higher CKA and lower Procrustes error are favorable. The figure includes all five shared-space models and uses every evaluated text layer.
    }
    \label{fig:app_paired_depth}
\end{figure}

We next evaluate text--video correspondence using linear CKA and held-out orthogonal Procrustes error. \Cref{fig:app_paired_depth} reports both measures for text and video layers aligned by relative depth on TUNA and VATEX. For PE-AV, each of the 22 text layers is paired with the video layer at the closest relative depth, so its four video layers are reused across multiple comparisons. 

The correspondence permutation analysis contains 32 comparisons: 4 models (PE-AV Small, PE-AV Large, X-CLIP, and Qwen3-VL) two retrieval tasks, and four prespecified relative-depth locations.
LCO is included in the layerwise analysis above but not in this permutation analysis.
For each comparison, 1,000 permutations shuffle the correspondence between text and video examples while leaving both sets of representation unchanged. 

The one-sided Monte Carlo value is
\[
p_{\mathrm{MC}}=\frac{1+b}{1001},
\]
where $b$ is the number of permutations at least as favorable as the observed value. The minimum attainable value is \(1/1001\).

\begin{figure}[t]
    \centering
    \includegraphics[width=\linewidth]{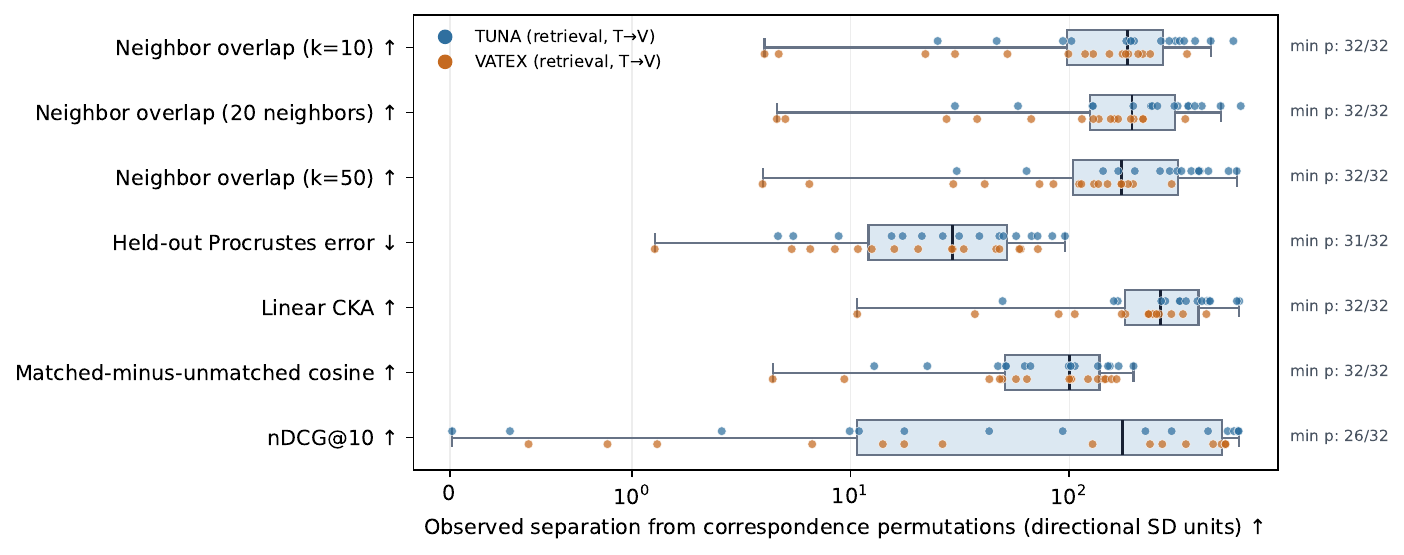}
    \caption{
        Separation between the observed text--video pairing and 1,000
        correspondence permutations across 32 model--task--depth
        comparisons.
    }
    \label{fig:app_pairing_controls}
\end{figure}

\Cref{fig:app_pairing_controls} reports separation from the permutation
distribution in directional standard-deviation units:
\[
z_{\mathrm{dir}}
=
s_g
\frac{g_{\mathrm{obs}}-\mu_{\mathrm{perm}}}
     {\sigma_{\mathrm{perm}}},
\]
where \(s_g=+1\) for measures for which larger values are favorable and
\(s_g=-1\) for held-out Procrustes error. Positive values favor the original
text--video pairing.

Held-out Procrustes error is more favorable than all 1,000 permutation values in 31 of 32 comparisons. The supplementary paired measures provide a similar check: linear CKA, matched-minus-unmatched cosine similarity, and neighbor overlap reach the minimum Monte Carlo value in all 32 comparisons. Retrieval itself is stricter. nDCG@10 beats all 1,000 permutations in 26 comparisons and satisfies \(p_{\mathrm{MC}}\leq0.05\) in 27. At the final evaluated depth, the paired measures and retrieval score all favor the observed correspondence.

\subsection{UCF101 zero-shot classification}
\label{app:ucf101_zero_shot_details}

For UCF101 zero-shot classification, we evaluate every text--video layer pair and the model-default pair for the five shared-space models. Four fixed prompt templates are evaluated individually and as a fixed ensemble, without selecting prompts based on evaluation accuracy. With the ensemble, the best layer pair achieves higher accuracy than the model-default pair in four models, with gains of up to $0.0082$. For X-CLIP, the best layer pair scores $0.0005$ lower than the model-default pair ({\Cref{tab:app_ucf101_zeroshot}).

\begin{table}[htb]
    \centering
    \caption{
        UCF101 zero-shot classification with four-prompt ensemble. Here, the best pair is selected from all text--video layer pairs.
    }
    \label{tab:app_ucf101_zeroshot}
    \small
    \setlength{\tabcolsep}{5pt}
    \renewcommand{\arraystretch}{1.08}
    \begin{tabular}{lcccc}
        \toprule
        Model &
        Best \((\ell_T,\ell_V)\) &
        Best accuracy &
        Model-default &
        Difference \\
        \midrule
        PE-AV Small & \(18/4\)  & 0.7449 & 0.7366 & +0.0082 \\
        PE-AV Large & \(21/4\)  & 0.8421 & 0.8390 & +0.0031 \\
        X-CLIP      & \(12/12\) & 0.7598 & 0.7603 & \(-0.0005\) \\
        Qwen3-VL    & \(27/27\) & 0.8724 & 0.8719 & +0.0005 \\
        LCO         & \(35/35\) & 0.8323 & 0.8302 & +0.0021 \\
        \bottomrule
    \end{tabular}
\end{table}

\section{Layerwise classification and clustering performance}
\label{app:classification_results}

This section reports the complete classification and clustering results
underlying \Cref{sec:results_geometry_performance} and \Cref{sec:results_layer_choice}. Every video layer is evaluated directly, with the model-default output evaluated separately.

\subsection{Primary classification and clustering results}
\label{app:primary_classification_results}

\Cref{tab:app_primary_layer_results} summarizes the all-layer results on
Breakfast classification and UCF101 classification and clustering. 

For each model--task comparison, \(\ell^\star\) is the layer with the highest mean score
over ten evaluator seeds. We define
\[
\Delta_{\mathrm{last}}
=
s^\star-s_L,
\qquad
\Delta_{\mathrm{default}}
=
s^\star-s_{\mathrm{default}},
\]
where \(s^\star\) is the score at \(\ell^\star\), \(s_L\) is the score at the
last layer, and \(s_{\mathrm{default}}\) is the score of the model-default
output. A positive difference favors the best layer. Differences are
computed from the unrounded scores.

\begin{table}[t]
    \centering
    \caption{
        Layerwise classification (cls.) and clustering (clust.) results on Breakfast and UCF101. Classification is measured by accuracy and clustering by V-measure. \(\ell_{best}\) is the layer with the highest mean score across ten evaluator seeds. \(s_{\mathrm{best}}\), \(s_{\mathrm{final}}\), and \(s_{\mathrm{default}}\) denote the corresponding ten-seed mean scores; the standard deviation shown with \(s_{\mathrm{best}}\) is computed at the fixed layer \(\ell_{\mathrm{best}}\). 
        }
    \label{tab:app_primary_layer_results}
    \scriptsize
    \setlength{\tabcolsep}{3.1pt}
    \renewcommand{\arraystretch}{1.07}
    \resizebox{\linewidth}{!}{
    \begin{tabular}{llccccccc}
        \toprule
        Model &
        Task &
        Layers &
        \(\ell_{best}\) &
        \(s_{best}\) &
        \(s_{final}\) &
        \(\Delta_{\mathrm{final}}\) &
        \(s_{\mathrm{default}}\) &
        \(\Delta_{\mathrm{default}}\) \\
        \midrule

        V-JEPA 2
        & Breakfast cls.
        & 24 & 22
        & \(0.3325 \pm 0.0243\)
        & 0.3110 & +0.0215
        & 0.3124 & +0.0201 \\

        V-JEPA 2
        & UCF101 cls.
        & 24 & 23
        & \(0.8129 \pm 0.0045\)
        & 0.8072 & +0.0058
        & 0.8062 & +0.0068 \\

        V-JEPA 2
        & UCF101 clust.
        & 24 & 23
        & \(0.7707 \pm 0.0040\)
        & 0.7702 & +0.0006
        & 0.7830 & \(-0.0123\) \\

        \midrule

        PE-AV Small
        & Breakfast cls.
        & 4 & 4
        & \(0.5185 \pm 0.0228\)
        & 0.5185 & +0.0000
        & 0.4529 & +0.0656 \\

        PE-AV Small
        & UCF101 cls.
        & 4 & 1
        & \(0.9338 \pm 0.0027\)
        & 0.9248 & +0.0090
        & 0.8759 & +0.0579 \\

        PE-AV Small
        & UCF101 clust.
        & 4 & 1
        & \(0.8283 \pm 0.0040\)
        & 0.7589 & +0.0694
        & 0.8508 & \(-0.0225\) \\

        \midrule

        Gemma 4
        & Breakfast cls.
        & 42 & 15
        & \(0.1263 \pm 0.0103\)
        & 0.1057 & +0.0206
        & 0.0912 & +0.0351 \\

        Gemma 4
        & UCF101 cls.
        & 42 & 14
        & \(0.8017 \pm 0.0062\)
        & 0.7076 & +0.0941
        & 0.1884 & +0.6132 \\

        Gemma 4
        & UCF101 clust.
        & 42 & 13
        & \(0.7665 \pm 0.0018\)
        & 0.5982 & +0.1683
        & 0.3457 & +0.4208 \\

        \midrule

        PE-AV Large
        & Breakfast cls.
        & 4 & 2
        & \(0.5295 \pm 0.0154\)
        & 0.5200 & +0.0095
        & 0.4741 & +0.0554 \\

        PE-AV Large
        & UCF101 cls.
        & 4 & 1
        & \(0.9275 \pm 0.0019\)
        & 0.8960 & +0.0315
        & 0.8991 & +0.0284 \\

        PE-AV Large
        & UCF101 clust.
        & 4 & 1
        & \(0.9075 \pm 0.0041\)
        & 0.8806 & +0.0268
        & 0.8776 & +0.0299 \\

        \midrule

        X-CLIP
        & Breakfast cls.
        & 12 & 12
        & \(0.1203 \pm 0.0127\)
        & 0.1203 & +0.0000
        & 0.2143 & \(-0.0940\) \\

        X-CLIP
        & UCF101 cls.
        & 12 & 12
        & \(0.8359 \pm 0.0037\)
        & 0.8359 & +0.0000
        & 0.9154 & \(-0.0795\) \\

        X-CLIP
        & UCF101 clust.
        & 12 & 10
        & \(0.8051 \pm 0.0035\)
        & 0.7960 & +0.0092
        & 0.9223 & \(-0.1172\) \\

        \midrule

        Qwen3-VL
        & Breakfast cls.
        & 28 & 28
        & \(0.4646 \pm 0.0197\)
        & 0.4646 & +0.0000
        & 0.6032 & \(-0.1386\) \\

        Qwen3-VL
        & UCF101 cls.
        & 28 & 28
        & \(0.9413 \pm 0.0026\)
        & 0.9413 & +0.0000
        & 0.9367 & +0.0047 \\

        Qwen3-VL
        & UCF101 clust.
        & 28 & 28
        & \(0.9258 \pm 0.0017\)
        & 0.9258 & +0.0000
        & 0.9486 & \(-0.0227\) \\

        \midrule

        LCO
        & Breakfast cls.
        & 36 & 36
        & \(0.4327 \pm 0.0149\)
        & 0.4327 & +0.0000
        & 0.5464 & \(-0.1136\) \\

        LCO
        & UCF101 cls.
        & 36 & 34
        & \(0.9268 \pm 0.0017\)
        & 0.9238 & +0.0030
        & 0.9273 & \(-0.0005\) \\

        LCO
        & UCF101 clust.
        & 36 & 35
        & \(0.9141 \pm 0.0024\)
        & 0.9079 & +0.0062
        & 0.9360 & \(-0.0219\) \\

        \bottomrule
    \end{tabular}}
\end{table}

\subsection{HMDB51 classification replication}
\label{app:hmdb51_results}

\begin{table}[t]
    \centering
    \caption{
        HMDB51 eight-shot classification across all evaluated video layers. \(\ell_{\mathrm{best}}\) is selected by mean accuracy across ten evaluator seeds. \(s_{\mathrm{best}}\), \(s_{\mathrm{final}}\), and \(s_{\mathrm{default}}\) are the corresponding mean accuracies. Positive differences favor the best raw layer.
    }
    \label{tab:app_hmdb51_results}
    \small
    \setlength{\tabcolsep}{4.2pt}
    \renewcommand{\arraystretch}{1.08}
    \begin{tabular}{lcccccc}
        \toprule
        Model &
        \(\ell_{best}\) &
        \(s_{best}\) &
        \(s_{final}\) &
        \(s_{\mathrm{default}}\) &
        \(\Delta_{\mathrm{final}}\) &
        \(\Delta_{\mathrm{default}}\) \\
        \midrule
        V-JEPA 2     & 23 & 0.4690 & 0.4682 & 0.4578 & +0.0009 & +0.0113 \\
        PE-AV Small  & 1  & 0.6284 & 0.5987 & 0.4247 & +0.0297 & +0.2036 \\
        Gemma 4      & 14 & 0.4724 & 0.3904 & 0.0891 & +0.0820 & +0.3833 \\
        PE-AV Large  & 2  & 0.6462 & 0.6237 & 0.5136 & +0.0225 & +0.1326 \\
        X-CLIP       & 12 & 0.4807 & 0.4807 & 0.6242 & +0.0000 & \(-0.1435\) \\
        Qwen3-VL     & 28 & 0.6224 & 0.6224 & 0.6602 & +0.0000 & \(-0.0378\) \\
        LCO           & 33 & 0.6006 & 0.5901 & 0.6466 & +0.0105 & \(-0.0460\) \\
        \bottomrule
    \end{tabular}
\end{table}

We repeat the eight-shot classification analysis on HMDB51 using the same layerwise evaluation protocol. This experiment provides a separate test of the intermediate-layer result, the results are detailed in \Cref{tab:app_hmdb51_results}.
We see that an intermediate layer outperforms the final layer in five of seven models and the model-default output in four. X-CLIP, Qwen3-VL, and LCO instead perform best with their model-default outputs.

\section{Geometry and downstream performance}
\label{app:results_geometry_performance}

This section gives the full model--task breakdown behind
\Cref{sec:results_geometry_performance}. The primary analysis contains 21 model–task comparisons: seven models on Breakfast classification, UCF101 classification, and UCF101 clustering.

\subsection{Geometry--performance associations}
\label{app:geometry_performance_associations}

\Cref{fig:app_geometry_associations} reports the within-model, within-task Spearman correlations between each geometric measure and downstream performance across layers. Downstream performance at each layer is averaged over ten evaluator seeds, while correlations are computed separately for each model–task comparison. 
Correlations are not reported for the four-layer PE-AV models and are marked \texttt{LR} (low resolution). \texttt{UD} denotes an undefined correlation because at least one ranked variable is degenerate.

\begin{figure}[t]
    \centering
    \includegraphics[width=0.92\linewidth]
    {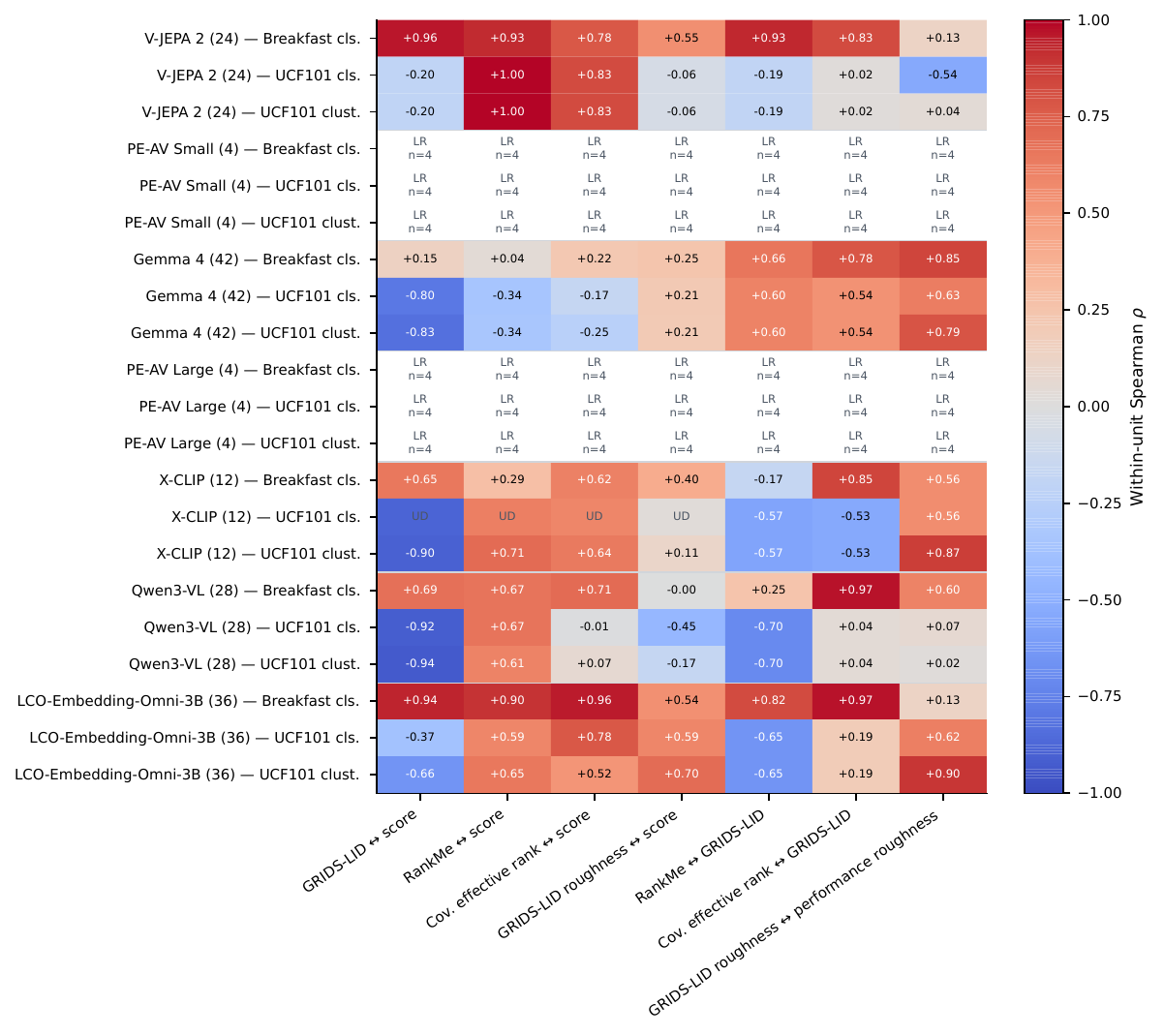}
    \caption{
        Within-model, within-task Spearman associations between GRIDS-LID, RankMe, standardized covariance effective rank, and
        downstream performance across layers. Task scores are ten-seed means. \textsc{lr} denotes a four-layer trajectory, for which we do not report a primary correlation; \textsc{ud} denotes an undefined association. Correlations use the \([-1,1]\) scale.
    }
    \label{fig:app_geometry_associations}
\end{figure}

The direction of the association varies across tasks. For LCO, the GRIDS-LID correlation changes from \(+0.94\) on Breakfast to \(-0.37\) on UCF101 classification and \(-0.66\) on UCF101 clustering. RankMe and standardized covariance effective rank are more often positively associated with performance, but are not consistent across models. For example, both are negative for Gemma 4 on the UCF101 tasks. 

\subsection{Layer-choice regret}
\label{app:layer_choice_regret_details}

For each model--task comparison, we select layers using minimum and maximum GRIDS-LID, maximum RankMe, and maximum standardized covariance effective rank. The final layer is included as a fixed baseline. The geometry-based selection is fixed across evaluator seeds, while the best-performing layer is determined separately for each seed. We compute regret for seeds \(0,\ldots,9\) and average it within each model-task comparison.

\begin{table}[t]
    \centering
    \caption{
        Layer-choice regret across the 21 primary classification and
        clustering comparisons. Regret uses the native \(0\)--\(1\)
        task-score scale; lower values are better. ``Zero'' counts
        comparisons with zero mean regret. Win/tie/loss compares each
        geometric choice with the last-layer baseline.
    }
    \label{tab:app_layer_choice_regret_summary}
    \small
    \setlength{\tabcolsep}{4.5pt}
    \renewcommand{\arraystretch}{1.08}
    \resizebox{\linewidth}{!}{
    \begin{tabular}{lrrrrrr}
        \toprule
        Layer choice &
        \(N\) &
        Mean &
        Median &
        Maximum &
        Zero &
        W/T/L vs.\ last \\
        \midrule
        Maximum RankMe
        & 21 & 0.0143 & 0.0081 & 0.0558 & 3 & 8/9/4 \\
        Last layer
        & 21 & 0.0249 & 0.0090 & 0.1694 & 4 & -- \\
        Maximum standardized covariance effective rank
        & 21 & 0.0399 & 0.0147 & 0.2172 & 3 & 10/3/8 \\
        Minimum GRIDS-LID (\(k=100\))
        & 21 & 0.0447 & 0.0063 & 0.2380 & 4 & 7/4/10 \\
        Maximum GRIDS-LID (\(k=100\))
        & 21 & 0.0997 & 0.0374 & 0.4734 & 0 & 3/6/12 \\
        \bottomrule
    \end{tabular}}
\end{table}

\Cref{tab:app_layer_choice_regret_summary} summarizes the 21 comparisons.
Maximum RankMe has the lowest mean regret, 0.0143, compared to 0.0249 for the final layer baseline. It improves on the final layer in eight comparison, ties in nine, and performs worse in four. Maximum standardized covariance effective rank improves on the final layer in ten comparisons, but its larger errors give a higher mean regret of 0.0399. 

Minimum GRIDS-LID has a low median regret of \(0.0063\), but a mean regret of 0.0447, reflecting several large errors. Maximum GRRIDS-LID performs worst overall, with a mean regret of 0.0997 and a maximum regret of 0.4734.

\Cref{fig:app_layer_choice_regret} shows that the GRIDS-LID errors depend strongly on the task. Maximum GRIDS-LID performs better on several Breakfast comparisons but incurs large regret on UCF101, partcularly for V-JEPA 2 and X-CLIP. Minimum GRIDS-LID shows the opposite pattern in several cases. Neither GRIDS-LID extremum therefore provides a consistent layer-selection rule across the evaluated tasks.

\begin{figure}[t]
    \centering
    \includegraphics[width=0.92\linewidth]
    {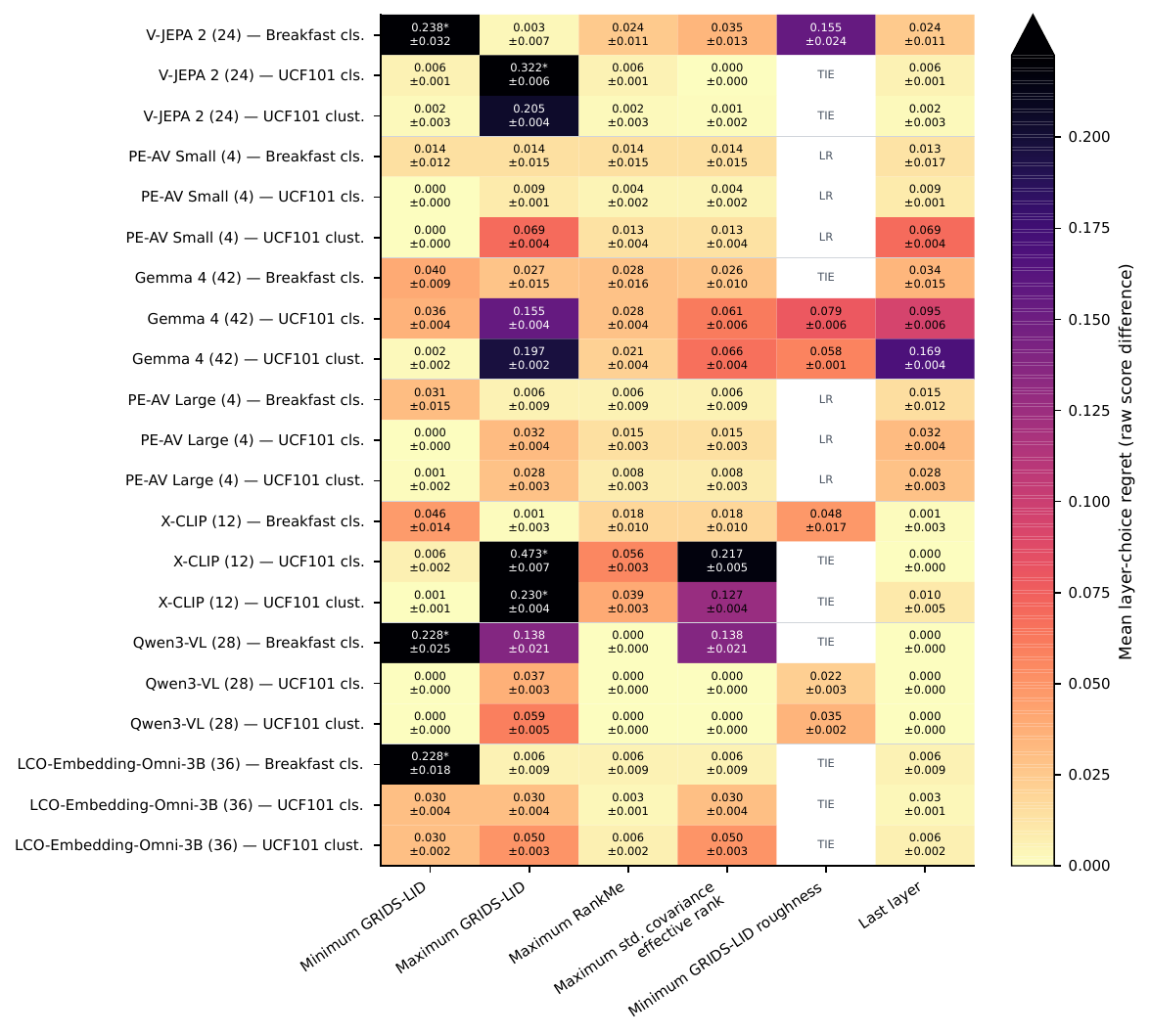}
    \caption{
        Layer-choice regret for each of the 21 primary classification
        and clustering comparisons. Each cell reports the mean and
        sample standard deviation over evaluator seeds \(0\)--\(9\) on
        the \(0\)--\(1\) task-score scale. Minimum and maximum
        GRIDS-LID are evaluated separately, and the last layer is a
        fixed baseline. Asterisks mark values above the displayed
        95th-percentile color cap.
    }
    \label{fig:app_layer_choice_regret}
\end{figure}

\clearpage

\end{document}